\documentclass[letterpaper, 10 pt, conference]{ieeeconf}  

\IEEEoverridecommandlockouts                              

\usepackage{hyperref} 
\usepackage{amsmath} 
\usepackage{amssymb}  
\usepackage{xcolor}
\usepackage{graphicx}
\usepackage{subcaption}
\usepackage{siunitx}
\usepackage[normalem]{ulem}
\useunder{\uline}{\ul}{}
\usepackage{xfrac} 
\renewcommand{\baselinestretch}{0.98} 
\title{\LARGE \bf
Impedance Optimization for Uncertain Contact Interactions Through Risk Sensitive Optimal Control
}

\author{Bilal Hammoud$^{1,2}$ , Majid Khadiv$^{2}$ and Ludovic Righetti$^{1,2}$
\thanks{*This work was supported by New York University, the European Union’s Horizon 2020 research and innovation program (grant agreement 780684 and European Research Council’s grant 637935) and the National Science Foundation (grants 1825993, 1932187 and 2026479).}
\DIFdelbegin 
\DIFdelend \DIFaddbegin \thanks{$^{1}$Tandon School of Engineering,
        New York University, USA}
\thanks{$^{2}$Max Planck Institute for Intelligent Systems, Tübingen, Germany {\tt\small first.last@tuebingen.mpg.de}}
}

\RequirePackage[normalem]{ulem} 
\RequirePackage{color}\definecolor{RED}{rgb}{1,0,0}\definecolor{BLUE}{rgb}{0,0,1} 
\providecommand{\DIFadd}[1]{{\protect\color{black}#1}}
\providecommand{\DIFaddbegin}{} 
\providecommand{\DIFaddend}{} 
\providecommand{\DIFdelbegin}{} 
\providecommand{\DIFdelend}{} 
\providecommand{\DIFdelFL}[1]{\DIFdel{#1}} 
\providecommand{\DIFaddbeginFL}{} 
\providecommand{\DIFaddendFL}{} 
\providecommand{\DIFdelbeginFL}{} 
\providecommand{\DIFdelendFL}{} 
\providecommand{\DIFadd}[1]{\texorpdfstring{\DIFaddtex{#1}}{#1}} 
\providecommand{\DIFdel}[1]{}
\newcommand{\DIFscaledelfig}{0.5}
\RequirePackage{settobox} 
\RequirePackage{letltxmacro} 
\newsavebox{\DIFdelgraphicsbox} 
\newlength{\DIFdelgraphicswidth} 
\newlength{\DIFdelgraphicsheight} 
\LetLtxMacro{\DIFOincludegraphics}{\includegraphics} 
\newcommand{\DIFaddincludegraphics}[2][]{{\color{blue}\fbox{\DIFOincludegraphics[#1]{#2}}}} 
\newcommand{\DIFdelincludegraphics}[2][]{
\sbox{\DIFdelgraphicsbox}{\DIFOincludegraphics[#1]{#2}}
\settoboxwidth{\DIFdelgraphicswidth}{\DIFdelgraphicsbox} 
\settoboxtotalheight{\DIFdelgraphicsheight}{\DIFdelgraphicsbox} 
\scalebox{\DIFscaledelfig}{
\parbox[b]{\DIFdelgraphicswidth}{\usebox{\DIFdelgraphicsbox}\\[-\baselineskip] \rule{\DIFdelgraphicswidth}{0em}}\llap{\resizebox{\DIFdelgraphicswidth}{\DIFdelgraphicsheight}{
\setlength{\unitlength}{\DIFdelgraphicswidth}
\begin{picture}(1,1)
\thicklines\linethickness{2pt} 
{\color[rgb]{1,0,0}\put(0,0){\framebox(1,1){}}}
{\color[rgb]{1,0,0}\put(0,0){\line( 1,1){1}}}
{\color[rgb]{1,0,0}\put(0,1){\line(1,-1){1}}}
\end{picture}
}\hspace*{3pt}}} 
} 
\LetLtxMacro{\DIFOaddbegin}{\DIFaddbegin} 
\LetLtxMacro{\DIFOaddend}{\DIFaddend} 
\LetLtxMacro{\DIFOdelbegin}{\DIFdelbegin} 
\LetLtxMacro{\DIFOdelend}{\DIFdelend} 
\DeclareRobustCommand{\DIFaddbegin}{\DIFOaddbegin \let\includegraphics\DIFaddincludegraphics} 
\DeclareRobustCommand{\DIFaddend}{\DIFOaddend \let\includegraphics\DIFOincludegraphics} 
\DeclareRobustCommand{\DIFdelbegin}{\DIFOdelbegin \let\includegraphics\DIFdelincludegraphics} 
\DeclareRobustCommand{\DIFdelend}{\DIFOaddend \let\includegraphics\DIFOincludegraphics} 
\LetLtxMacro{\DIFOaddbeginFL}{\DIFaddbeginFL} 
\LetLtxMacro{\DIFOaddendFL}{\DIFaddendFL} 
\LetLtxMacro{\DIFOdelbeginFL}{\DIFdelbeginFL} 
\LetLtxMacro{\DIFOdelendFL}{\DIFdelendFL} 
\DeclareRobustCommand{\DIFaddbeginFL}{\DIFOaddbeginFL \let\includegraphics\DIFaddincludegraphics} 
\DeclareRobustCommand{\DIFaddendFL}{\DIFOaddendFL \let\includegraphics\DIFOincludegraphics} 
\DeclareRobustCommand{\DIFdelbeginFL}{\DIFOdelbeginFL \let\includegraphics\DIFdelincludegraphics} 
\DeclareRobustCommand{\DIFdelendFL}{\DIFOaddendFL \let\includegraphics\DIFOincludegraphics} 

\begin{document}

\maketitle
\thispagestyle{empty}
\pagestyle{empty}

\begin{abstract}
This paper addresses the problem of computing optimal impedance schedules
for legged locomotion tasks involving complex contact interactions.
We formulate the problem of impedance regulation as a trade-off between
disturbance rejection and measurement uncertainty.
We extend a stochastic optimal control algorithm known as Risk Sensitive Control to take into account measurement uncertainty and propose a formal way to include such uncertainty for unknown contact locations.
The approach can efficiently generate optimal state and control trajectories along with local feedback control gains, i.e. impedance schedules. Extensive simulations demonstrate the capabilities of the approach in generating meaningful stiffness and damping modulation patterns
before and after contact interaction.
For example, contact forces are reduced during early contacts, damping increases to anticipate a high impact event and tracking is automatically traded-off for increased stability.
In particular, we show a significant
improvement in performance during jumping and trotting tasks with a simulated quadruped robot.
\end{abstract}

\section{Introduction}

State of the art locomotion controllers include a model predictive control scheme that computes trajectories of some reduced model. This model predictive scheme is then realized through a pre-designed impedance controller or a QP based inverse dynamics solver. These strategies have proven to be successful in completely structured and controlled environments. However, current robot control strategies still lack the ability to reason about uncertainty in the environment. High stiffness feedback controllers are usually used to track precise trajectories. This approach is usually limiting for a robot in multi-contact scenarios where the robot depends on intermittent contact interactions to move itself or some object around. Contact interactions increase the complexity of the control design problem. A stiff controller will counteract an unpredicted contact by increasing the control input to ensure tracking, which might generate high contact forces that destabilize the system. On the other hand, excessive compliance could lead to large deviations from the desired task. Studies from the field of neuroscience suggest that human beings modulate their impedance during contact interactions~\cite{burdet2001central,lee2016summary}. Other studies suggest that sensorimotor commands are the result of an optimal feedback control mechanism that controls a trade-off between accuracy and stability~\cite{liu2007evidence}. \DIFdelbegin \DIFdel{More recent studies~\mbox{
\cite{nagengast2011risk,braun2011risk,crevecoeur2014beyond} }\hspace{0pt}
suggest }\DIFdelend \DIFaddbegin \DIFadd{A more recent study~\mbox{
\cite{nagengast2011risk} }\hspace{0pt}
suggests }\DIFaddend that this impedance "sweet spot" is a result of reasoning not only about the desired task, but also the uncertainties present during contact interactions. 

Impedance control for robotics was introduced \DIFdelbegin \DIFdel{in \mbox{
\cite{hogan1985impedancepart1} }\hspace{0pt}
and a }\DIFdelend \DIFaddbegin \DIFadd{by Hogan~\mbox{
\cite{hogan1985impedancepart1} }\hspace{0pt}
and the }\DIFaddend controller was derived \DIFdelbegin \DIFdel{in \mbox{
\cite{hogan1985impedancepart2}}\hspace{0pt}
. Later, }\DIFdelend \DIFaddbegin \DIFadd{later~\mbox{
\cite{hogan1985impedancepart2}}\hspace{0pt}
. Hogan }\DIFaddend \cite{hogan1987stable} demonstrated experimentally that this controller is capable of stabilizing contact interactions if proper impedance parameters \DIFdelbegin \DIFdel{are }\DIFdelend \DIFaddbegin \DIFadd{were to be }\DIFaddend chosen. Following the results presented by Hogan, Park\DIFaddbegin \DIFadd{~}\DIFaddend \cite{park2001impedance} used a similar approach to design a bipedal walking controller. The results show a biped capable \DIFdelbegin \DIFdel{to walk }\DIFdelend \DIFaddbegin \DIFadd{of walking }\DIFaddend on uneven terrain. \DIFdelbegin \DIFdel{A similar approach was used for a quadruped trotting gait in~\mbox{
\cite{park2012impedance}}\hspace{0pt}
. Another application of impedance control was presented in~\mbox{
\cite{lee2014dynamics} }\hspace{0pt}
for the MIT Cheetah, where an impedance controller is shown to be stable at various speeds of a periodic trotting gait. In~\mbox{
\cite{boaventura2015model} }\hspace{0pt}
the authors provided a thorough study on the impedance controller of the HyQ quadruped robot using hydraulic actuators. }\DIFdelend The results achieved through impedance control have proven to be superior for control strategies involving contact interactions\DIFaddbegin \DIFadd{~\mbox{
\cite{lee2014dynamics}}\hspace{0pt}
}\DIFaddend . However, \DIFdelbegin \DIFdel{all these }\DIFdelend \DIFaddbegin \DIFadd{the mentioned }\DIFaddend approaches design the impedance \DIFdelbegin \DIFdel{schedule }\DIFdelend \DIFaddbegin \DIFadd{schedules }\DIFaddend through an exhaustive trial and error process. It remains an open question \DIFaddbegin \DIFadd{on }\DIFaddend how to systematically optimize impedance profiles for robotic tasks involving contact interactions. 

Optimal feedback control theory has many promising aspects that could help approach this problem. \DIFdelbegin \DIFdel{An optimal control formulation results in motions from an abstract task formulation, namely the cost function along with feedback gains to stabilize the motion, hence impedance profiles.  This aligns well with the suggested concepts in~\mbox{
\cite{liu2007evidence,nagengast2011risk}}\hspace{0pt}
. An }\DIFdelend \DIFaddbegin \DIFadd{Mayne~\mbox{
\cite{mayne1966second} }\hspace{0pt}
introduced an }\DIFaddend algorithm that computes local quadratic approximations of both the dynamics and the cost functions and then iteratively solves the nonlinear optimal control problem\DIFdelbegin \DIFdel{was introduced by Mayne~\mbox{
\cite{mayne1966second}}\hspace{0pt}
}\DIFdelend . This algorithm is commonly known as the Differential Dynamic Programming Algorithm (\textit{DDP}). It  has the advantage of providing an optimal control trajectory together with local feedback controller. Many variations of \textit{DDP} appeared later in the literature\DIFdelbegin \DIFdel{Li~\mbox{
\cite{li2004iterative}}\hspace{0pt}
, Sideris~\mbox{
\cite{sideris2005efficient}}\hspace{0pt}
, Tassa~\mbox{
\cite{tassa2012synthesis} }\hspace{0pt}
and ~\mbox{
\cite{tassa2014control}}\hspace{0pt}
}\DIFdelend \DIFaddbegin \DIFadd{~\mbox{
\cite{li2004iterative, sideris2005efficient, tassa2012synthesis,tassa2014control,grandia2019frequency, grandia2019feedback}}\hspace{0pt}
}\DIFaddend . 
\DIFdelbegin \DIFdel{In \mbox{
\cite{grandia2019frequency} }\hspace{0pt}
the issue of control bandwidth in optimal control is addressed. The work in \mbox{
\cite{grandia2019feedback} }\hspace{0pt}
extended the previous results to optimize a reduced model of the quadruped robot ANYmal. These algorithms }\DIFdelend 
\DIFaddbegin \DIFadd{However, all the mentioned variations }\DIFaddend are deterministic in nature and favor tracking over stability, making them prone to failure in situations where tracking cannot be perfectly achieved, and in attempting to do so, the controller can destabilize the system, uncertain contact interactions being a clear example. 

Todorov~\cite{todorov2005generalized} added multiplicative process noise to the optimal control problem violating the certainty equivalence \DIFdelbegin \DIFdel{principles}\DIFdelend \DIFaddbegin \DIFadd{principle}\DIFaddend . This led to control policies that are dependent on the process noise.  Li~\cite{li2007iterative} derived similar results for partially observable systems with control constraints resulting in control policies that are a function of both process and measurement uncertainties. Another method to break the certainty equivalence principle is achieved through an exponential transformation of the cost function, this was first introduced by Jacobson\DIFdelbegin \DIFdel{in}\DIFdelend ~\cite{jacobson1973optimal} for linear systems and later extended for nonlinear optimal control \DIFdelbegin \DIFdel{using a DDP-like algorithm by Farshidianin}\DIFdelend \DIFaddbegin \DIFadd{by Farshidian}\DIFaddend ~\cite{farshidian2015risk}. This formally synthesizes a controller that could obtain risk neutral, risk sensitive or risk seeking behaviors depending on the parameterization of the role of the uncertainty in the cost function.  Medina~\cite{medina2018considering} used the exponential cost transformation with process noise to perform manipulation tasks through a model predictive control scheme. The exponential transformation was extended to accommodate for measurement uncertainties in~\cite{ponton2020effects} obtaining a risk sensitive optimal control algorithm that accounts for higher order statistics in both process and measurement models making it a suitable framework for designing feedback \DIFdelbegin \DIFdel{controls that reason about the }\DIFdelend \DIFaddbegin \DIFadd{controllers that can }\DIFaddend trade-off \DIFdelbegin \DIFdel{between }\DIFdelend disturbance rejection and measurement uncertainty. \DIFdelbegin \DIFdel{However, the }\DIFdelend \DIFaddbegin \DIFadd{The }\DIFaddend approach was only tested on toy problems and never \DIFdelbegin \DIFdel{analyzed to achieve }\DIFdelend \DIFaddbegin \DIFadd{used for }\DIFaddend more complex robotic tasks. 

\DIFdelbegin \DIFdel{In this paper , we build }\DIFdelend \DIFaddbegin \DIFadd{This paper builds }\DIFaddend on the ideas of \DIFaddbegin \DIFadd{Ponton~}\DIFaddend \cite{ponton2020effects} to propose a systematic \DIFdelbegin \DIFdel{manner to compute }\DIFdelend \DIFaddbegin \DIFadd{method for computing }\DIFaddend impedance schedules for legged robots. We extend the algorithm to work with hard contact transitions and introduce a way \DIFdelbegin \DIFdel{of incorporating }\DIFdelend \DIFaddbegin \DIFadd{to incorporate }\DIFaddend contact measurement uncertainty into the whole-body optimal control formulation. This results in systematically optimized impedance profiles that exhibit desirable stiffness and damping patterns to handle uncertain, high impact \DIFdelbegin \DIFdel{, }\DIFdelend \DIFaddbegin \DIFadd{and }\DIFaddend contact transitions. Extensive numerical simulations demonstrate the properties of the approach when compared to usual DDP algorithms and other measurement noise models. In particular, we show a significant increase in performance for hard impacts \DIFdelbegin \DIFdel{during jumping and for }\DIFdelend \DIFaddbegin \DIFadd{for jumping and }\DIFaddend trotting over uneven terrains.

\section{Background}\label{sec:background}
\DIFdelbegin \DIFdel{In this section , we provide }\DIFdelend \DIFaddbegin \DIFadd{This section provides }\DIFaddend background on the robot and contact models, risk-sensitive stochastic optimal control and its
extension to include measurement uncertainty.
\subsection{Multi-Contact Robot Dynamics}\label{sec:muticontact_ocp}
The dynamics of a legged robot in contact with its environment \DIFdelbegin \DIFdel{can be described by the follow equation 
}\DIFdelend \DIFaddbegin \DIFadd{is described using the following equation 
}\vspace{-0.05cm}
 \DIFaddend \begin{align}\DIFdelbegin 
\DIFdel{M}
\DIFdel{q}
\DIFdel{+ h}
\DIFdel{q,v}
\DIFdelend \DIFaddbegin \label{eqn:first_order_dynamics}
     \underbrace{\frac{d}{dt} \begin{bmatrix} q \\ v \end{bmatrix}  }_{\DIFadd{\dot{x}}\DIFaddend} = \DIFdelbegin \DIFdel{S^T \tau + J}
\DIFdel{q}
\DIFdel{^T \lambda_{ext} 
}\DIFdelend \DIFaddbegin \underbrace{ \begin{bmatrix} v \\ M^{-1} (\tau - h + J^T \lambda_{ext}) \end{bmatrix} }_{\DIFadd{f(x, u)}
 }\DIFaddend \end{align}

where \DIFdelbegin \DIFdel{$q\in R^{nq}$ }\DIFdelend \DIFaddbegin \DIFadd{$q\in SE(3) \times R^{nj}$ }\DIFaddend is the vector of generalized coordinates \DIFaddbegin \DIFadd{with $nj$ being the number of robot joints}\DIFaddend , $v\in R^{nv}$ is the vector of generalized velocities \DIFdelbegin \DIFdel{, }\DIFdelend \DIFaddbegin \DIFadd{with $nv = nj+6$. }\DIFaddend $M\left(q\right)$ is the inertial matrix, $h\left(q,v\right)$ is the vector combining the nonlinear terms such as Coriolis acceleration and gravity, $S^T$ is the selection matrix mapping the controls to the actuated degrees of freedom, $\lambda_{ext}$ is the vector of contact forces and $J\left(q\right)$ is the contact Jacobian. The notation indicating the dependence on $q$ and $v$ will be omitted for the remainder of the text. \DIFaddbegin \DIFadd{Let $x^T=\left[q^T, v^T\right]$  define }\DIFaddend the state vector\DIFdelbegin \DIFdel{as  $x^T=\left[q^T, v^T\right]\in \mathcal{X}$, }\DIFdelend \DIFaddbegin \DIFadd{, hence the 
 discrete time state transitions become
}\vspace{-0.05cm}
 \begin{align}\DIFadd{\label{eqn:state_integration}
    x_{t+1} = f_t\left(x_t,u_t\right) = x_t \oplus \underbrace{\left(\delta t . f(x, u)\right)}_{dx_t}  
}\end{align}
\DIFadd{Then $dx_t$ represents }\DIFaddend the \DIFdelbegin \DIFdel{set of all possible states. Then the }\DIFdelend change in the state vector during a time interval $\delta t$ \DIFaddbegin \DIFadd{and }\DIFaddend $\oplus$ handles the Lie group composition operation for the base orientation.
 \DIFdelbegin 

\DIFdelend 
 \subsection{Rigid Contact Model}
While different contact models can be chosen to compute the contact forces $\lambda_{ext}$ \DIFdelbegin \DIFdel{\mbox{
\cite{stewart2000rigid}}\hspace{0pt}
, we will use }\DIFdelend \DIFaddbegin \DIFadd{\mbox{
\cite{gilardi2002literature}}\hspace{0pt}
, }\DIFaddend a rigid contact model \DIFaddbegin \DIFadd{is chosen }\DIFaddend for the optimal control computation\DIFaddbegin \DIFadd{~\mbox{
\cite{righetti2013optimal}}\hspace{0pt}
}\DIFaddend . Let $p$, $\dot{p}$ and $\Ddot{p}$ denote any contact point position, velocity and acceleration respectively\DIFdelbegin \DIFdel{. During }\DIFdelend \DIFaddbegin \DIFadd{, then during }\DIFaddend an active contact phase, the rigid contact assumption can be stated as
\begin{align}
    \dot{p} \DIFdelbegin 
\DIFdelend = J v = 0  \DIFdelbegin 
\DIFdelend \DIFaddbegin \quad \textrm{\DIFadd{and}} \quad \DIFaddend \Ddot{p} \DIFdelbegin 
\DIFdelend = \dot{J}v + J \dot{v} = 0 \DIFdelbegin 
\DIFdelend \DIFaddbegin \DIFadd{.}\label{eq:rigid_contact_constraints}
\DIFaddend \end{align}
\noindent In order to resolve the contact forces \DIFdelbegin \DIFdel{for }\DIFdelend \DIFaddbegin \DIFadd{that guarantee the no-motion constraints of }\DIFaddend all active contacts, the robot dynamics~\DIFdelbegin 
\DIFdelend \DIFaddbegin \eqref{eqn:first_order_dynamics} \DIFaddend is projected to the contact space using $\left(J M^{-1}J^T\right)^{-1} J M^{-1} = \Lambda  J M^{-1} $ to result in 
\begin{align}
 \lambda_{ext} =   -\Lambda \dot{J}v + \Lambda J M^{-1} h - \Lambda J M^{-1} S^T \tau 
\end{align}
\noindent Once $\lambda_{ext}$ is obtained, the motion vector \DIFdelbegin \DIFdel{~}
\DIFdel{can be constructed }\DIFdelend \DIFaddbegin \DIFadd{$dx_t$ can be computed }\DIFaddend and the state vector $x_{t+1}$ can be obtained from~\eqref{eqn:state_integration}.

\subsection{Risk Sensitive Optimal Control}
We \DIFdelbegin \DIFdel{are interested in }\DIFdelend \DIFaddbegin \DIFadd{consider the }\DIFaddend stochastic optimal control \DIFdelbegin \DIFdel{approaches that
can explicitly handle uncertainty. In particular, we use
a }\DIFdelend \DIFaddbegin \DIFadd{approach known as Risk Sensitive Control to explicitly reason about uncertainty. A }\DIFaddend nonlinear iterative risk sensitive optimal control formulation~\DIFdelbegin \DIFdel{\mbox{
\cite{dvijotham2012unifying, farshidian2015risk, ponton2020effects} }\hspace{0pt}
which enables to explicitly take }\DIFdelend \DIFaddbegin \DIFadd{\mbox{
\cite{dvijotham2011unifying, farshidian2015risk, ponton2020effects} }\hspace{0pt}
explicitly takes }\DIFaddend into account the distribution of uncertainty while being numerically efficient \DIFdelbegin \DIFdel{.
The algorithm is similar to what can be obtained using DDP~\mbox{
\cite{tassa2014control, mastalli2019crocoddyl} }\hspace{0pt}
while reasoning about the higher order statistics of the problem. }\DIFdelend \DIFaddbegin \DIFadd{for nonlinear problems. }\DIFaddend Consider the following dynamics \DIFdelbegin \DIFdel{expressed }\DIFdelend \DIFaddbegin \DIFadd{written }\DIFaddend as a nonlinear stochastic difference equation
\begin{align}
    x_{t+1} = f\DIFaddbegin \DIFadd{_t}\DIFaddend \left(x_t,u_t\right) + F\left(x_t,u_t\right) \omega_t \label{Eq:dynamics_sde}
\end{align}
\DIFaddbegin \noindent \DIFaddend where $\omega_t \sim \mathcal{N}\left(0,\Omega_t\right)$ is the process noise \DIFdelbegin \DIFdel{(i.e. it }\DIFdelend \DIFaddbegin \DIFadd{which }\DIFaddend accounts for unmodeled disturbances \DIFdelbegin \DIFdel{) }\DIFdelend and $F(x_t, u_t)$ maps the noise to the full \DIFdelbegin \DIFdel{dynamics. A typical optimal control problem will optimize }\DIFdelend \DIFaddbegin \DIFadd{state. Consider }\DIFaddend an objective function of the form
\begin{align}
    \mathcal{L}\left(\DIFdelbegin \DIFdel{x}\DIFdelend \DIFaddbegin \DIFadd{\mathcal{X}}\DIFaddend ,\DIFdelbegin \DIFdel{\pi}
\DIFdel{x}\DIFdelend \DIFaddbegin \DIFadd{\mathcal{U}}\DIFaddend \right) \DIFdelbegin 
\DIFdelend = l_T\left(x_T\right) + \sum_{0}^{T-1} l_t\left(x_t,\DIFdelbegin \DIFdel{\pi}
\DIFdel{x}\DIFdelend \DIFaddbegin \DIFadd{u}\DIFaddend _t\right) \DIFdelbegin 
\DIFdelend \label{Eq:functional} 
\end{align}
where \DIFaddbegin \DIFadd{$\mathcal{X}=[x_0, ..., x_T]$ and $\mathcal{U}=[u_0,...,u_{T-1}]$ denote the state and control trajectories respectively, }\DIFaddend $l_T\left(x_T\right)$ is the \DIFdelbegin \DIFdel{cost at the terminal time $T$, and  $l_t\left(x_t,\pi\left(x_t\right)\right)$ }\DIFdelend \DIFaddbegin \DIFadd{terminal cost, and  $l_t\left(x_t,u_t\right)$ }\DIFaddend is the cost at time $t$\DIFdelbegin \DIFdel{along the horizon. In order to include the higher order statistics into the problem, }\DIFdelend \DIFaddbegin \DIFadd{. Typical optimal control approaches minimize the expectation of the objective function, however, }\DIFaddend risk sensitive optimal control \DIFdelbegin \DIFdel{minimizes the following }\DIFdelend \DIFaddbegin \DIFadd{instead minimizes the expectation of the exponential }\DIFaddend transformation of the objective\DIFaddbegin \DIFadd{. This affords the consideration of the higher-order statistics of the cost
}\DIFaddend %
\begin{align}
     \mathcal{J}^{*}\  &= \min\DIFdelbegin \DIFdel{_{\pi\left(x_t\right)} }\DIFdelend _{\DIFaddbegin \DIFadd{\mathcal{X},\mathcal{U}} \DIFaddend} \   \mathbb{E}\left[\DIFdelbegin \DIFdel{\exp{\left(\sigma \mathcal{L}\left(x,\pi\right)\right)}}\DIFdelend \DIFaddbegin \DIFadd{\exp{\left(\sigma \mathcal{L}\left(\mathcal{X},\mathcal{U}\right)\right)}}\DIFaddend \right]  \label{Eq:stoch_ocp_cost}
\end{align}
\DIFdelbegin \DIFdel{where $\mathcal{J}^{*}$ is the optimal risk sensitive cost and }\DIFdelend 
\DIFaddbegin \DIFadd{where }\DIFaddend $\sigma$ is the \DIFdelbegin \DIFdel{risk sensitive parameter}\DIFdelend \DIFaddbegin \DIFadd{sensitivity scalar}\DIFaddend . Farshidian~\cite{farshidian2015risk} proved that the cumulant generating function of $\mathcal{J}$ can be expressed as 
\begin{align}
    \frac{1}{\sigma} \log{\mathcal{J}} = \mathbb{E}\left[\mathcal{L}\right]  + \frac{\sigma}{2} \mu_2\left[\mathcal{L}\right] + \frac{\sigma^2}{6} \mu_3\left[\mathcal{L}\right]+ ... 
\end{align}
where $\mu_i\left[\mathcal{L}\right]$ is the $i$'th moment of the random variable $\mathcal{L}$. The risk sensitivity parameter $\sigma$ then provides a tool to control the contribution of the higher order moments on the cost. \DIFaddbegin \DIFadd{When $\sigma<0$, the control is risk-seeking and
higher cost variances will be preferred. When $\sigma>0$,
the control is }\DIFaddend risk-averse \DIFaddbegin \DIFadd{since a high variance of the
cost distribution will be more penalized. When $\sigma\to 0$,
the problem reduces to a normal (risk-neutral) optimal control problem and only the expectation of the objective is minimized.} 

\DIFadd{The problem can be solved globally with Ricatti-like equations
for linear dynamics and a quadratic cost function~\mbox{
\cite{jacobson1973optimal}}\hspace{0pt}
, the solution being a linear feedback
controller. However, finding a global minimizer $\mathcal{X}^*, \mathcal{U}^*$ for the cost function~}\eqref{Eq:stoch_ocp_cost} \DIFadd{is generally intractable for nonlinear dynamics and non quadratic costs. The method presented in this paper computes locally optimal solutions through iterative linearizations of the dynamics and a quadratic }\DIFaddend approximation
of the \DIFdelbegin \DIFdel{system dynamics along }\DIFdelend \DIFaddbegin \DIFadd{objective functions, a common technique used in deriving iterative nonlinear optimal control algorithms~\mbox{
\cite{mayne1966second, tassa2014control, farshidian2015risk, mastalli2020crocoddyl}}\hspace{0pt}
. The local deviations from }\DIFaddend nominal state and control trajectories, \DIFdelbegin \DIFdel{then finding improvements to the deviations~}
\DIFdel{along the state and control trajectories iteratively. }\DIFdelend \DIFaddbegin \DIFadd{denoted by superscript $n$, are written as
}\begin{align}
    \DIFadd{\delta x_t = x_t \ominus x^n_t, \quad \quad \delta u_t = u_t - u^n_t\label{eqn:deviations}
}\end{align}
\DIFaddend %
\DIFaddbegin \DIFadd{with $\ominus$ representing the suitable difference on the state manifold. The system dynamics can be linearized in terms of the deviations as   
}\vspace{-0.05cm}
\DIFaddend \begin{align}
    \delta x \DIFaddbegin \DIFadd{_{t+1} }\DIFaddend = \DIFdelbegin \DIFdel{x}\DIFdelend \DIFaddbegin \DIFadd{A}\DIFaddend _t \DIFaddbegin \DIFadd{\delta }\DIFaddend x_t \DIFaddbegin \DIFadd{+ B_t }\DIFaddend \delta u_t  \DIFaddbegin \DIFadd{+ C}\DIFaddend _t \DIFaddbegin \DIFadd{\omega}\DIFaddend _t  \DIFaddbegin \label{eq:linearized_dynamics}
\DIFaddend \end{align}
\DIFaddbegin \DIFadd{where $A_t$, $B_t$ and $C_t$ are the respective linearization of $f_t(\delta x_t, \delta u_t)$ and $F_t(\delta x_t, \delta u_t)$ with respect to }\DIFaddend the \DIFaddbegin \DIFadd{state and control terms. Similarly a quadratic approximation of the cost function $l_t(x_t, u_t)$ can be obtained. For a linear dynamics and a quadratic cost under the risk sensitive exponential transformation, the value function takes the form of an exponential with a quadratic argument in the state~\mbox{
\cite{jacobson1973optimal} 
}\hspace{0pt}
}\vspace{-0.05cm}
\begin{align}
    \DIFadd{V\left(\delta x_t\right) }&\DIFadd{= \exp{ \Bigg\lbrace \sigma \left(\frac{1}{2}\delta x_t^T S_t \delta x_t + \delta x_t^T s_t + \Bar{s}_t\right) \Bigg\rbrace} \label{eq:value_approximation}
}\end{align} 
\DIFadd{Importantly, the value function here holds a completely different form than that of DDP and the principle of optimality needs to be written in multiplicative form as 
}\vspace{-0.05cm}
\DIFaddend \begin{align}
    \DIFaddbegin \DIFadd{V}\left(\DIFaddend \delta \DIFdelbegin \DIFdel{u}\DIFdelend \DIFaddbegin \DIFadd{x}\DIFaddend _t\DIFaddbegin \right) &\DIFaddend = \DIFdelbegin \DIFdel{K_t }\DIFdelend \DIFaddbegin \DIFadd{\min_{\delta u_t} }\Big\lbrace \DIFadd{\exp{\left[\sigma  l_t\left(\delta x_t,\delta u_t\right)\right]}\  \mathbb{E}}\left[ \DIFadd{V}\left(\DIFaddend \delta x\DIFdelbegin \DIFdel{_t + k_t
}\DIFdelend \DIFaddbegin \DIFadd{_{t+1}}\right) \right] \Big\rbrace \label{eq:value_recursion}
\DIFaddend \end{align}
\DIFdelbegin \DIFdel{where }\DIFdelend \DIFaddbegin 

\DIFadd{In order to obtain the solution, the expectation of the value function at time $t+1$ must be computed. Finally the value function at time $t$ is approximated by computing $S_t, s_t$ and $\Bar{s}_t$ along with $\delta u_t$ recursively backward in time. Since $\delta u_t$ is the argument minimizing a quadratic in $\delta x_t$, then the locally optimal controller $\delta u_t = K_t \delta x_t + k_t$ is linear in the state deviations. Here, }\DIFaddend $K_t$ \DIFdelbegin \DIFdel{are }\DIFdelend \DIFaddbegin \DIFadd{is }\DIFaddend the error feedback gains \DIFdelbegin \DIFdel{and }\DIFdelend \DIFaddbegin \DIFadd{while }\DIFaddend $k_t$ \DIFdelbegin \DIFdel{the feedforward commands. Notably, the }\DIFdelend \DIFaddbegin \DIFadd{is the feedforward command. The solution of the recursive risk sensitive value function is discussed in details in~\mbox{
\cite{jacobson1973optimal, farshidian2015risk}}\hspace{0pt}
. The optimized deviations are used to iteratively improve the solution until convergence is reached. It is important to note that the }\DIFaddend control law explicitly incorporates the covariance of the noise distribution in $K_t$ and $k_t$\DIFaddbegin \DIFadd{~}\DIFaddend \cite{farshidian2015risk}.

\subsection{Including Measurement Uncertainty} 
The formulation presented above \DIFdelbegin \DIFdel{constructs the optimal control solutions for a problem with process noise.
 }\DIFdelend \DIFaddbegin \DIFadd{does not integrate the uncertainty coming from a nonlinear measurement model
}\begin{align}
    \DIFadd{y_{t+1} = g\left(x_t,u_t\right) + H\left(x_t,u_t\right)\gamma_t\label{Eq:measurement_model}
 }\end{align}
 \DIFadd{where $\gamma_t \sim \mathcal{N}\left(0, \Gamma_t\right)$ is the measurement noise.
 Measurement uncertainty is crucial for our approach as we need to
 take into account the uncertainty about contact locations.
}\DIFaddend Speyer~\cite{speyer1974optimization} incorporated measurement noise into the risk sensitive formulation by introducing a state vector that grows at each time step to include the entire history of the states. Recently, Ponton~\cite{ponton2020effects} suggested that this could be avoided by \DIFdelbegin \DIFdel{extending }\DIFdelend \DIFaddbegin \DIFadd{augmenting }\DIFaddend the linearized system dynamics with that of an Extended Kalman Filter (EKF). Assuming an observable system, an EKF can be used to compute the estimate deviations $\delta \hat{x}_t$ along the nominal trajectory at each iteration. \DIFaddbegin \DIFadd{Let }\DIFaddend $F_t$ and $D_t$ \DIFdelbegin \DIFdel{are the local }\DIFdelend \DIFaddbegin \DIFadd{be the }\DIFaddend linear approximations of \DIFdelbegin \DIFdel{$f\left(x_t,u_t\right)$, $F\left(x_t,u_t\right)$, $h\left(x_t,u_t\right)$ }\DIFdelend \DIFaddbegin \DIFadd{$g\left(x_t,u_t\right)$ }\DIFaddend and $H\left(x_t,u_t\right)$ from~\DIFdelbegin 
\DIFdel{and~}\DIFdelend \eqref{Eq:measurement_model} respectively. \DIFdelbegin \DIFdel{At each iteration, }\DIFdelend \DIFaddbegin \DIFadd{Then we can compute }\DIFaddend the Kalman gains $G_t$ \DIFdelbegin \DIFdel{are computed along the nominal trajectory, then the optimal control law is optimized by iterating over modified backward Riccati equations. The optimal feedback gains $K_t$ explicitly include the covariance of both  process and measurement noises, i.e. the control law depends on a trade-off between external disturbances and measurement uncertainty}\DIFdelend \DIFaddbegin \DIFadd{during the forward pass of the iterative algorithm (i.e.
when updating the nominal state and control trajectories) and use them to define an augmented linear system, which includes both the real, $\delta x_t$, and estimated, $\delta \hat{x}_t$, states of the system. With the extended state $\delta \Tilde{x}_t = [\delta x_t, \delta \Hat{x}_t]$, the augmented
system dynamics becomes
}\begin{align}
    \DIFadd{\begin{bmatrix} \delta x_{t+1} \\ \delta \hat{x}_{t+1} \end{bmatrix} =}& \DIFadd{\begin{bmatrix}A_t\delta x_t + B_t \delta u_t \\ A_t\delta x_t + B_t \delta u_t + G_t F_t \left(\delta x_t - \delta \hat{x}_t\right) \end{bmatrix} \nonumber }\\ &\DIFadd{+ \begin{bmatrix} C_t & 0  \\ 0 & G_t D_t \end{bmatrix} \begin{bmatrix}\omega_t \\ \gamma_t \end{bmatrix} \label{eq:extended_dynamics}
}\end{align}
\DIFadd{This augmented linear dynamical system is used, in place of }\eqref{eq:linearized_dynamics}\DIFadd{, to solve the risk sensitive problem. The resulting locally optimal control policy is then 
}\vspace{-0.05cm}
\begin{align}
    \DIFadd{\delta u_t^* = k_t + \begin{bmatrix} K_t^x & K_t^{\hat{x}} \end{bmatrix}\delta \Tilde{x}_t
}\end{align} 
 \DIFadd{As the true state deviations $\delta x_t$ are unknown, then following Li~\mbox{
\cite{li2007iterative} }\hspace{0pt}
and Ponton~\mbox{
\cite{ponton2020effects}}\hspace{0pt}
, we take the expectation of $\delta u_t^*$ with $\delta x_t$ conditioned on $\delta \hat{x}_t$ to give the optimal policy
}\vspace{-0.05cm}
\begin{align}
    \DIFadd{\mathbb{E}_{\delta x_t \vert \delta \hat{x}_t} \left[\delta u_t^*\right]  = k_t + \left(K_t^x + K_t^{\hat{x}} \right)\delta \hat{x}_t
}\end{align}
\DIFadd{Detailed derivations of the complete recursive algorithm follow the  work of Ponton~\mbox{
\cite{ponton2020effects} }\hspace{0pt}
and can be found together with our software implementation in our open-source repository}\footnote{\url{https://github.com/machines-in-motion/risk\_sensitive\_control}}\DIFadd{. It is also important to note that the backward recursion has the same complexity as that of DDP.  The  only  added  complexity  is  the  computation  of  the estimator gains in the forward pass. The convergence of iterative risk sensitive solvers is studied in the work of Roulet~\mbox{
\cite{roulet2019convergence}}\hspace{0pt}
}\DIFaddend .


\DIFdelbegin 
\DIFdelend 
\DIFaddbegin \section{Multi-Contact Risk Sensitive Control}
Now we detail how we use the risk-sensitive optimal control algorithm including measurement uncertainties to optimize motions and impedance schedules for legged robots.
\subsection{Including Multiple Contact Switching}
\DIFdelbegin \DIFdel{Since we assume a }\DIFdelend 
\DIFaddbegin \DIFadd{For the }\DIFaddend rigid contact model\DIFdelbegin \DIFdel{, we need to define the switching dynamics to handle discontinuous contact transitions. We use
}\DIFdelend \DIFaddbegin \DIFadd{~}\eqref{eq:rigid_contact_constraints}\DIFadd{, a contact switch causes a change in the dimensions of the contact Jacobian $J$. Given }\DIFaddend a predefined contact sequence and timing \DIFdelbegin \DIFdel{for contact switching and compute
}\DIFdelend \DIFaddbegin \DIFadd{we can define the switching in the overall system dynamics denoted $f_n$ where the subscript $n$ indicates a predefined contact phase. The contact switching sequence is obtained with }\DIFaddend an initial guess for the open-loop optimal trajectory using an existing kino-dynamic optimizer \cite{ponton2018time}. \DIFdelbegin 

\DIFdel{We also align the contact transitions }\DIFdelend \DIFaddbegin \DIFadd{The contact transitions are also aligned }\DIFaddend with the collocation points such that \DIFdelbegin \DIFdel{we have }\DIFdelend the objective function and dynamics of the problem \DIFdelbegin \DIFdel{in the form 
}\DIFdelend \DIFaddbegin \DIFadd{can be written as
}\DIFaddend \begin{align}
     \min_{\delta X,\delta U} &\DIFdelbegin 
\DIFdel{l_T}
\DIFdel{\delta x_T}
\DIFdel{+ \sum_{n=0}^{N} \sum_{t=0}^{T} l_t}
\DIFdel{\delta x_t, \delta u_t}
\DIFdelend \DIFaddbegin \DIFadd{\mathbb{E}}\left[\DIFadd{\exp{\left(\sigma l_T\left(\delta x_T\right) + \sigma \sum_{n=0}^{N} \sum_{t=0}^{T} l_t\left(\delta x_t, \delta u_t\right) \right)}}\right] \DIFaddend \\
     \DIFaddbegin &\DIFaddend s.t. \DIFdelbegin 
\DIFdelend \quad \delta x_{t+1} = f_n \left(\delta x_t, \delta  u_t\right)
\end{align}
where \DIFaddbegin \DIFadd{$\delta \mathcal{X}$ and $\delta \mathcal{U}$ are the trajectories of the state and control deviations respectively.  }\DIFaddend $N$ is the number of contact switches along the trajectory\DIFaddbegin \DIFadd{, }\DIFaddend and $T$ is the horizon of each phase. This procedure avoids non-smooth switches in the contacts by solving the optimal control problem on multiple smooth intervals. \DIFdelbegin \DIFdel{We enforce consistency }\DIFdelend \DIFaddbegin \DIFadd{Consistency }\DIFaddend at the transition between the switching intervals \DIFdelbegin \DIFdel{through additional cost terms (i. e. soft constraints). }\DIFdelend \DIFaddbegin \DIFadd{is ensured through the forward integration of the trajectory along with high penalty on tracking the switching states. }\DIFaddend In particular, \DIFdelbegin \DIFdel{we enforce linearized
friction cone constraints and smooth contact force transitions.
}\DIFdelend We compute analytical derivatives for all the quantities, including the contacts, using the Pinocchio \DIFdelbegin \DIFdel{library \mbox{
\cite{carpentier2019pinocchio}}\hspace{0pt}
}\DIFdelend \DIFaddbegin \DIFadd{and Crocoddyl libraries~\mbox{
\cite{carpentier2019pinocchio, mastalli2020crocoddyl}}\hspace{0pt}
}\DIFaddend .

\subsection{Error State Kalman Filter on Smooth Manifolds}
 \DIFaddbegin \DIFadd{It is desired to design an EKF to augment the linearized dynamics of the robot at each time step, hence, the same dynamical model $f_n$ is used, i.e. at each time step, the predefined contact sequence defines the dynamical model of choice. Let the discrete time EKF of the state deviations $\delta \Hat{x}_t$ take the form of the second row of~}\eqref{eq:extended_dynamics}\DIFadd{, then following~\mbox{
\cite{ponton2020effects} }\hspace{0pt}
the error dynamics $\delta e_{t+1} = \delta x_{t+1} - \delta \hat{x}_{t+1}$ is given by 
}\vspace{-0.05 cm}
\begin{align}
    \DIFadd{\delta e_{t+1} = \left(A_t - G_t F_t\right)\delta e_t + C_t \omega_t - G_t D_t\gamma_t
}\end{align}\DIFaddend 
\DIFaddbegin 

\DIFadd{This is the simplest estimator of choice and can be replaced with any other estimator as long as it can be written locally as a linear dynamical system. The optimal estimation gains are then given by 
}\begin{align}
    \DIFadd{G_t = A_t \Sigma_t F_t^T\left(F_t \Sigma_t F_t^T + D_t \Gamma_t D_t^T\right)^{-1}
}\end{align}
\DIFaddend %
where \DIFdelbegin \DIFdel{$\ominus$ represents the difference between the two states $\hat{x}$ and  $x$ with the proper Lie group operations~\mbox{
\cite{sola2018micro}}\hspace{0pt}
. The propagation of}\DIFdelend \DIFaddbegin \DIFadd{$\Sigma_t$ is the  error dynamics covariance. As legged robots have a free floating base modeled as an $SE(3)$ element, }\DIFaddend the \DIFaddbegin \DIFadd{proper Lie group operations must be utilized during the propagation of the covariance of the local error dynamics. }\DIFaddend A full treatment of EKFs on \DIFdelbegin \DIFdel{lie }\DIFdelend \DIFaddbegin \DIFadd{Lie }\DIFaddend groups can be found in~\DIFdelbegin \DIFdel{\mbox{
\cite{sola2018micro,sola2017quaternion}}\hspace{0pt}
.
}\DIFdelend \DIFaddbegin \DIFadd{\mbox{
\cite{sola2017quaternion}}\hspace{0pt}
.
}\DIFaddend

\DIFaddbegin

\DIFaddend \subsection{Uncertainty in Contact Interactions}\label{sec:contact_uncertainty}
\begin{figure}[h]
\centering
\captionsetup{justification=centering}
\DIFdelbeginFL 
\DIFdelendFL \DIFaddbeginFL \includegraphics[width=.4\linewidth]{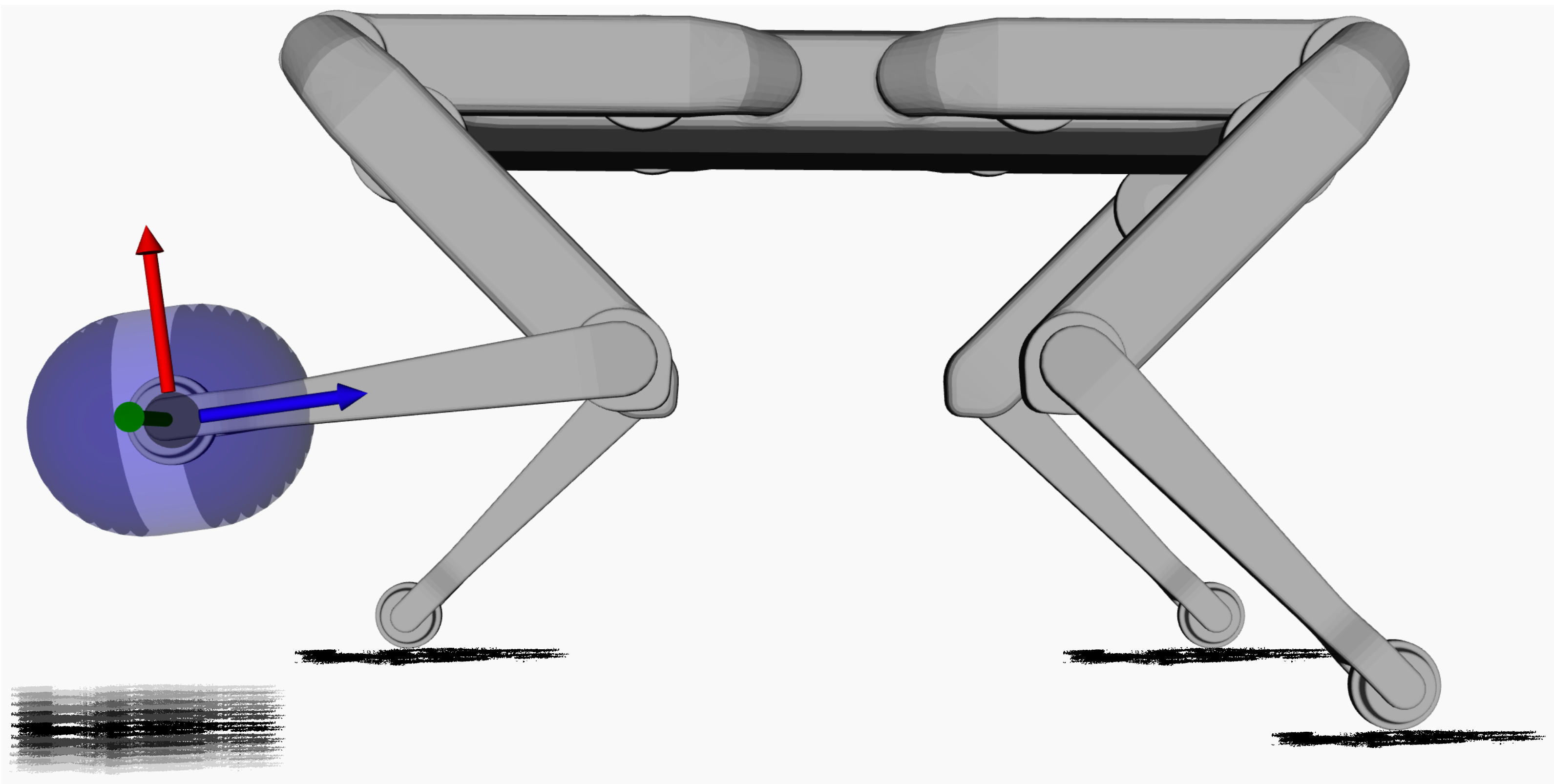}
\DIFaddendFL \caption{Uncertainty in Contact Interactions}
\label{fig:solo_contact_uncertainty}
\DIFaddbeginFL \vspace{-0.5cm}
\DIFaddendFL \end{figure}

Consider a robot during a dynamic \DIFdelbegin \DIFdel{multi-contact scenario such as the quadruped }\DIFdelend \DIFaddbegin \DIFadd{contact scenario as depicted }\DIFaddend in Fig.~\ref{fig:solo_contact_uncertainty}. A \DIFdelbegin \DIFdel{robot }\DIFdelend control algorithm computes \DIFdelbegin \DIFdel{control }\DIFdelend \DIFaddbegin \DIFadd{actuation }\DIFaddend torques based on the perception of the robot's own states along with its surrounding environment. Both the perception of the robot environment and its states are susceptible to sensor noise and perception errors, introducing uncertainty in the measurements. 
We propose to model the contact uncertainty as an uncertainty at the tip of the swing foot $\Gamma_c$ as shown in the ellipsoid in Fig.~\ref{fig:solo_contact_uncertainty}. This has the advantage \DIFdelbegin \DIFdel{that we do not }\DIFdelend \DIFaddbegin \DIFadd{of avoiding the }\DIFaddend need to keep track of the environment model \DIFdelbegin \DIFdel{nor }\DIFdelend \DIFaddbegin \DIFadd{or }\DIFaddend the next contact location in \DIFdelbegin \DIFdel{our }\DIFdelend \DIFaddbegin \DIFadd{the }\DIFaddend state vector. We then map this uncertainty back to the space where the full state covariance matrix $\Gamma_{fs}$ is defined. Adding both \DIFdelbegin \DIFdel{covariance matrices then }\DIFdelend \DIFaddbegin \DIFadd{covariances }\DIFaddend results in the total uncertainty in the contact interaction. 

A deviation in the swinging end-effector of the robot can be linearly mapped to a deviation in its state vector through 
\begin{align} \label{Eq:endeffector_deviations}
     \begin{bmatrix} \delta p \\ \delta \dot{p} \end{bmatrix} = \underbrace{\begin{bmatrix} J_s & 0 \\ \Dot{J}_s & J_s \end{bmatrix}}_{A_s} \begin{bmatrix} \delta q \\ \delta v \end{bmatrix}
\end{align}
\DIFaddbegin \DIFadd{where $J_s$ is the jacobian of the swinging foot. }\DIFaddend Then the minimum norm change in the state vector corresponding to a change in the end-effector is given  by
\begin{align}
    \begin{bmatrix} \delta q \\ \delta v \end{bmatrix}   =A^{\dagger}_s \begin{bmatrix} \delta p \\ \delta \dot{p} \end{bmatrix} \label{Eq:contact_uncertainty_model}
\end{align}
where $A^{\dagger}_s = A^T_s \left(A_s.A^T_s\right)^{-1}$ is the Moore-Penrose inverse of $A_s$. Different norms could be chosen using a weighted inverse if desired. Now that the deviations of the end-effector are in the same vector space as the full state errors, it is possible to add the noise resulting from the robot states such as joint encoders along with the contact uncertainty.  However, a deviation in the swing foot might induce a deviation in the feet that are actively in contact and break our rigid constraint assumption in~\DIFdelbegin 
\DIFdel{and~}
\DIFdelend \DIFaddbegin \eqref{eq:rigid_contact_constraints}\DIFaddend . To avoid inconsistency, the deviations of the swing foot must be projected to the null space of the feet actively in contact \DIFaddbegin \DIFadd{using the null space projector $P_c$.
}\DIFaddend %
\DIFaddbegin \DIFadd{The final form of the deviations transformation can then be written as 
}\DIFaddend \begin{align}
    \DIFdelbegin \DIFdel{P_c }\DIFdelend \DIFaddbegin \begin{bmatrix} \delta q \\ \delta v \end{bmatrix}   \DIFaddend =\DIFdelbegin \DIFdel{I - }\DIFdelend \DIFaddbegin \underbrace{\left(I - A_c^\dagger A_c\right)}_{\DIFadd{P_c} }\DIFaddend A\DIFdelbegin \DIFdel{_c^\dagger A_c }
\DIFdelend \DIFaddbegin \DIFadd{_s^{\dagger} }\begin{bmatrix} \delta p \\ \delta \dot{p} \end{bmatrix} \label{Eq:contact_uncertainty_model_constrained}
\DIFaddend \end{align}
where $A_c$ is similar in structure to $A_s$ however containing the Jacobians of the active contacts described in~\DIFaddbegin \eqref{eq:rigid_contact_constraints}\DIFadd{. 
This transformation maps }\DIFaddend the mean of the deviations in a certain end-effector frame to that of the full state vector of the robot. Since \DIFdelbegin \DIFdel{the noise was assumed to be }\DIFdelend \DIFaddbegin \DIFadd{our noise model is }\DIFaddend Gaussian, the covariance matrix can also be transformed using the affine property of Gaussians. The total covariance of the state vector becomes 
\begin{align}
    \Gamma = \Gamma_{fs} + P_c A^{\dagger}\DIFaddbegin \DIFadd{_s }\DIFaddend \Gamma_{c} A^{\dagger^T}\DIFaddbegin \DIFadd{_s }\DIFaddend P_c^T\label{eq:covariance_proj}
\end{align}
In summary, we define Gaussian noise models both in end-effector and state space, with \DIFdelbegin \DIFdel{covariance $\Gamma_{fs}$ and }\DIFdelend \DIFaddbegin \DIFadd{respective covariance }\DIFaddend $\Gamma_{c}$ \DIFdelbegin \DIFdel{respectively and }\DIFdelend \DIFaddbegin \DIFadd{and $\Gamma_{fs}$. We }\DIFaddend combine them using \eqref{eq:covariance_proj}. \DIFdelbegin \DIFdel{It will enable us to explicitly change the uncertainty associated to the swinging feet before }\DIFdelend \DIFaddbegin \DIFadd{This approach allows to modulate the at the swing feet just before and after }\DIFaddend contact.



\section{Simulations Results}\label{sec:simulation}
To demonstrate the capabilities of the proposed method for controlling multi-contact interactions under contact uncertainty, we present three different simulation experiments using an accurate model of our open-source quadruped robot Solo~\cite{grimminger2020open}. 
In the first experiment, we study the effect of different measurement noise models on the computed impedance profiles and resulting impact forces when encountering an unexpected contact. With the second experiment, we explore how the trade-off between stiffness and damping changes when using our risk-sensitive control approach when compared to standard DDP methods. Finally, in the third experiment, we systematically quantify how the stability of the system is favored relative to the accuracy of the controller in the risk sensitive case during locomotion tasks.

The DDP controller used as a baseline in the presented experiments is from~\DIFdelbegin \DIFdel{\mbox{
\cite{mastalli2019crocoddyl}}\hspace{0pt}
}\DIFdelend \DIFaddbegin \DIFadd{\mbox{
\cite{mastalli2020crocoddyl}}\hspace{0pt}
}\DIFaddend . The kino-dynamic optimizer described in~\cite{ponton2018time} is used to generate reference trajectories around which both iterative controllers DDP and Risk Sensitive are initialized.  
%
A linear spring damper contact model is used for the simulations with an explicit Euler integration scheme \cite{hammoud_2020_contact} at a time step \DIFdelbegin \DIFdel{$\delta t_{sim} = 1.e{-4}\si{s}$ }\DIFdelend \DIFaddbegin \DIFadd{$\delta t_{sim} = \num{1e-4}\si{s}$ }\DIFaddend with a spring stiffness parameter of \DIFdelbegin \DIFdel{$k=1.e{+5}\si{N\per m}$ }\DIFdelend \DIFaddbegin \DIFadd{$k=\SI{1e5}{N\per m}$ }\DIFaddend and a spring damping parameter of \DIFdelbegin \DIFdel{$b=3.e{+2}\si{Ns\per m}$}\DIFdelend \DIFaddbegin \DIFadd{$b=\num{3e2}\si{Ns\per m}$}\DIFaddend . The coefficient of static friction used in simulation is $\mu = 0.7$. The simulated feedback control frequency runs at $1\si{kHz}$ and the discretization step for the optimal control problems is set to \DIFdelbegin \DIFdel{$\delta t_{opt} = 1.e{-2}\si{s}$. }\DIFdelend \DIFaddbegin \DIFadd{$\delta t_{opt} = \SI{1e-2}{s}$. Moreover, the legs of the robot will be referred to as $FL,FR,HL,HR$ denoting FrontLeft, FrontRight, HindLeft and HindRight respectively. Each leg of the robot in turn consists of three joints as it can be seen in Fig.~\ref{fig:solo_contact_uncertainty}. 
}\DIFaddend

\DIFaddbegin

\DIFaddend For all the experiments, the same cost function, weights and reference trajectories are used for both DDP and Risk Sensitive Control. All the planning and control is designed for perfectly flat floor in all three experiments. This results in the same whole body trajectories $x_t$ and the same feedforward torque control profiles $\tau$ for both DDP and Risk Sensitive control \DIFaddbegin \DIFadd{in each experiment}\DIFaddend . The sensitivity parameter is set to $\sigma = 10$ for the Risk Sensitive solver, leaving the uncertainty models as the only variable in the experiments. The only differences in the optimized plans are the impedance profiles (i.e. the feedback gains $K_t$).  

\DIFaddbegin \DIFadd{For all the experiments presented, the uncertainty parameters can be described as follows. The process noise $\Omega_t = diag(\num{1e-6})$ through out the whole experiment. The full state measurement noise  $\Gamma_{fs}=diag(\num{5e-3})$ for all the experiments.  $\Gamma_c = 0 $ during any lift off phase of the foot. For the landing phase $\Gamma_c$ is increased to $\num{5e-4}$ for the diagonal elements corresponding to $p$ and $\num{1e-4}$ for the diagonal elements corresponding to $\Dot{p}$. 
}

\DIFaddend 
\subsection{Effect of Noise Models on Impedance Regulation}
In this experiment, the task is to swing a single leg forward with a maximum height of $\SI{10}{cm}$ and a step length of $\SI{8}{cm}$ similar to what is shown in Fig.~\ref{fig:solo_contact_uncertainty}. A $\SI{3}{cm}$ high block is added at the next contact location to simulate an unpredicted contact of \DIFdelbegin \DIFdel{$~ 9.2\%$ }\DIFdelend \DIFaddbegin \DIFadd{$9.2\%$ }\DIFaddend of the total leg length. The contact with the block occurs at $t = \SI{0.43}{s}$ whereas the contact with flat ground was planned for $t=\SI{0.55}{s}$. 

\begin{figure}[h]
\vspace{0.2cm}
\centering
\captionsetup{justification=centering}
\subfloat[][Position error norm]{
\includegraphics[width=0.45\linewidth]{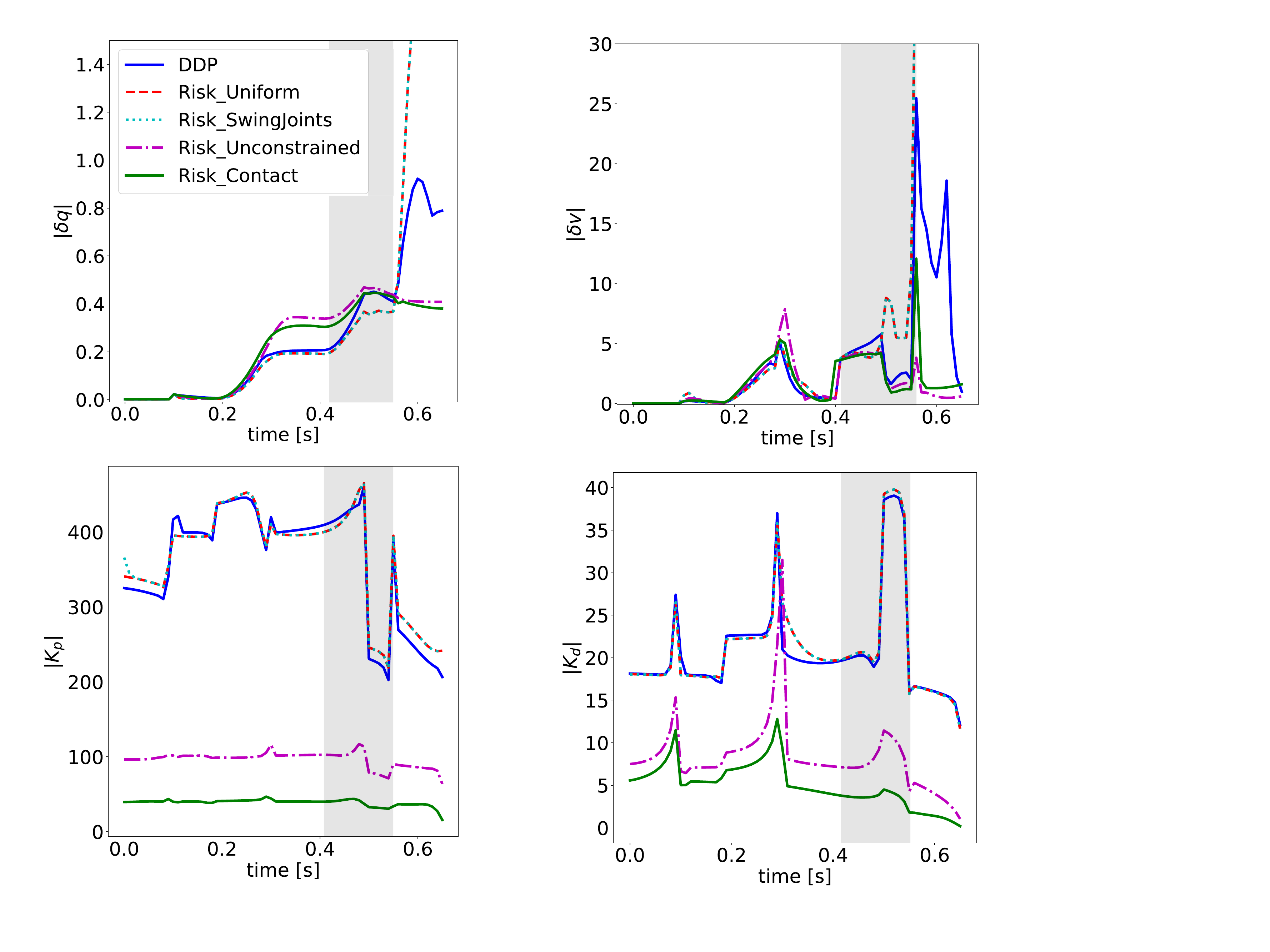}
\label{fig:exp01_q_error}}
\subfloat[][Velocity error norm]{
\includegraphics[width=0.45\linewidth]{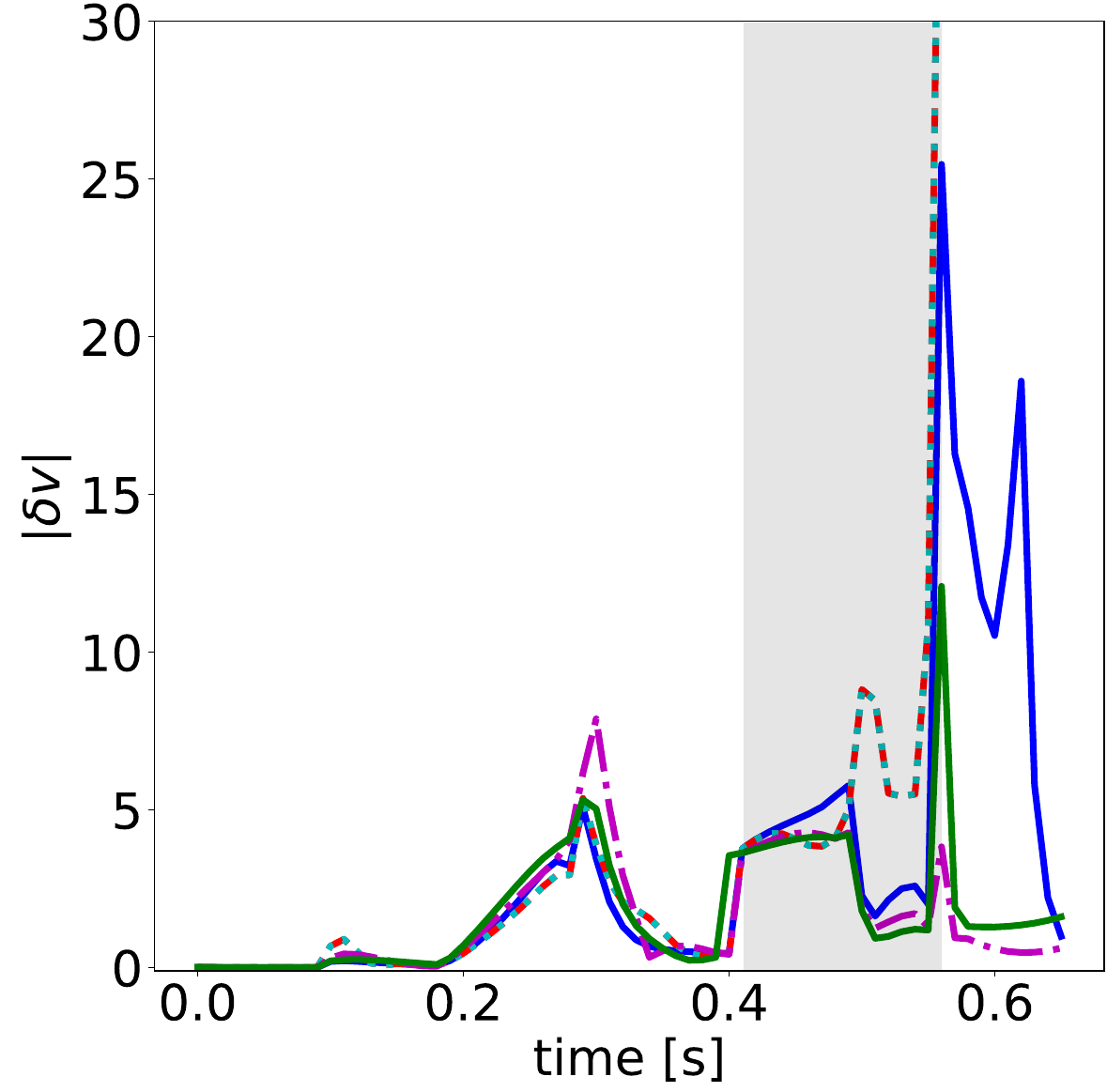}
\label{fig:exp01_v_error}} \\
\subfloat[][Stiffness norm]{
\includegraphics[width=0.45\linewidth]{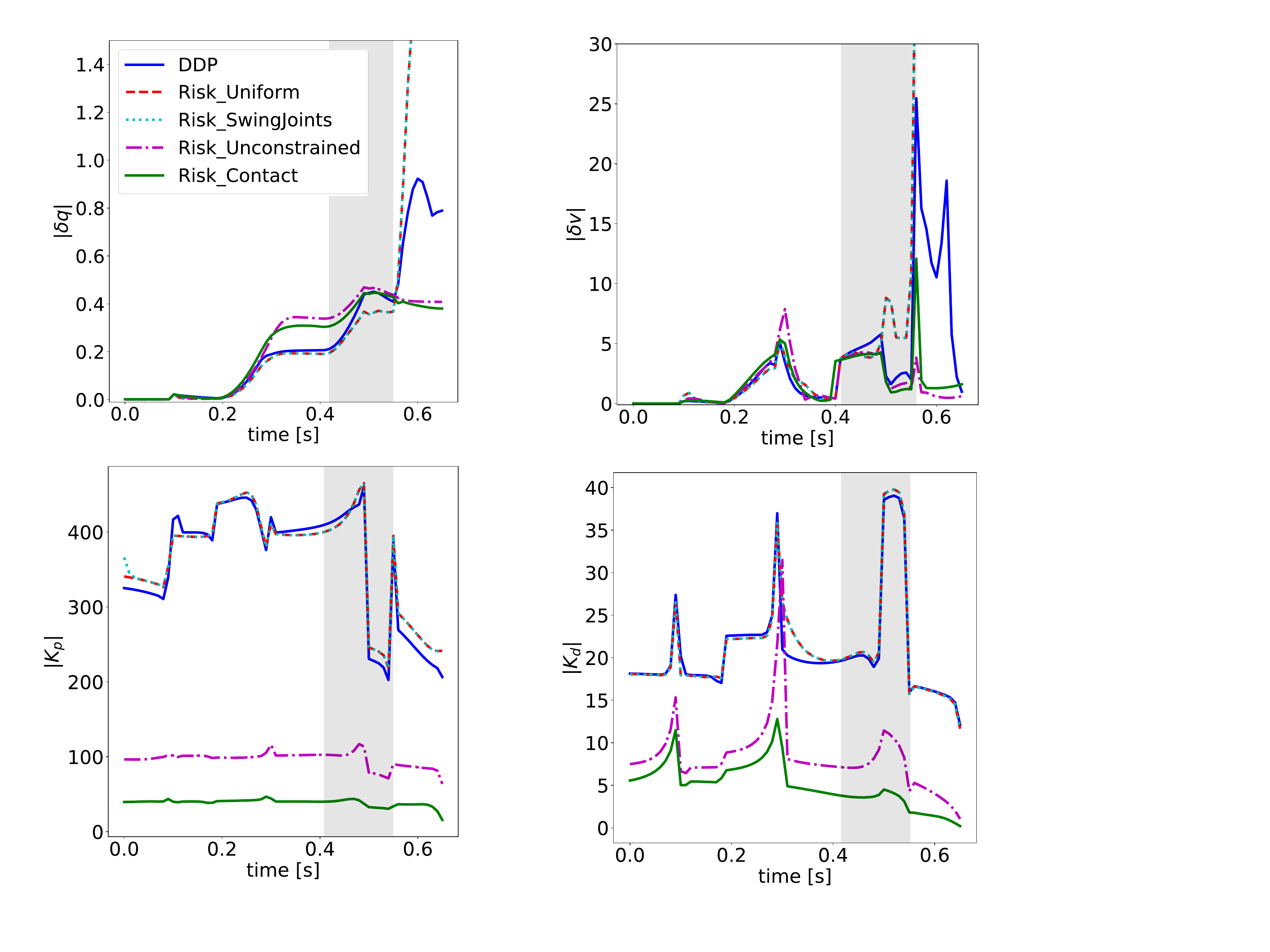}
\label{fig:exp01_kp}}
\subfloat[][Damping norm]{
\includegraphics[width=0.45\linewidth]{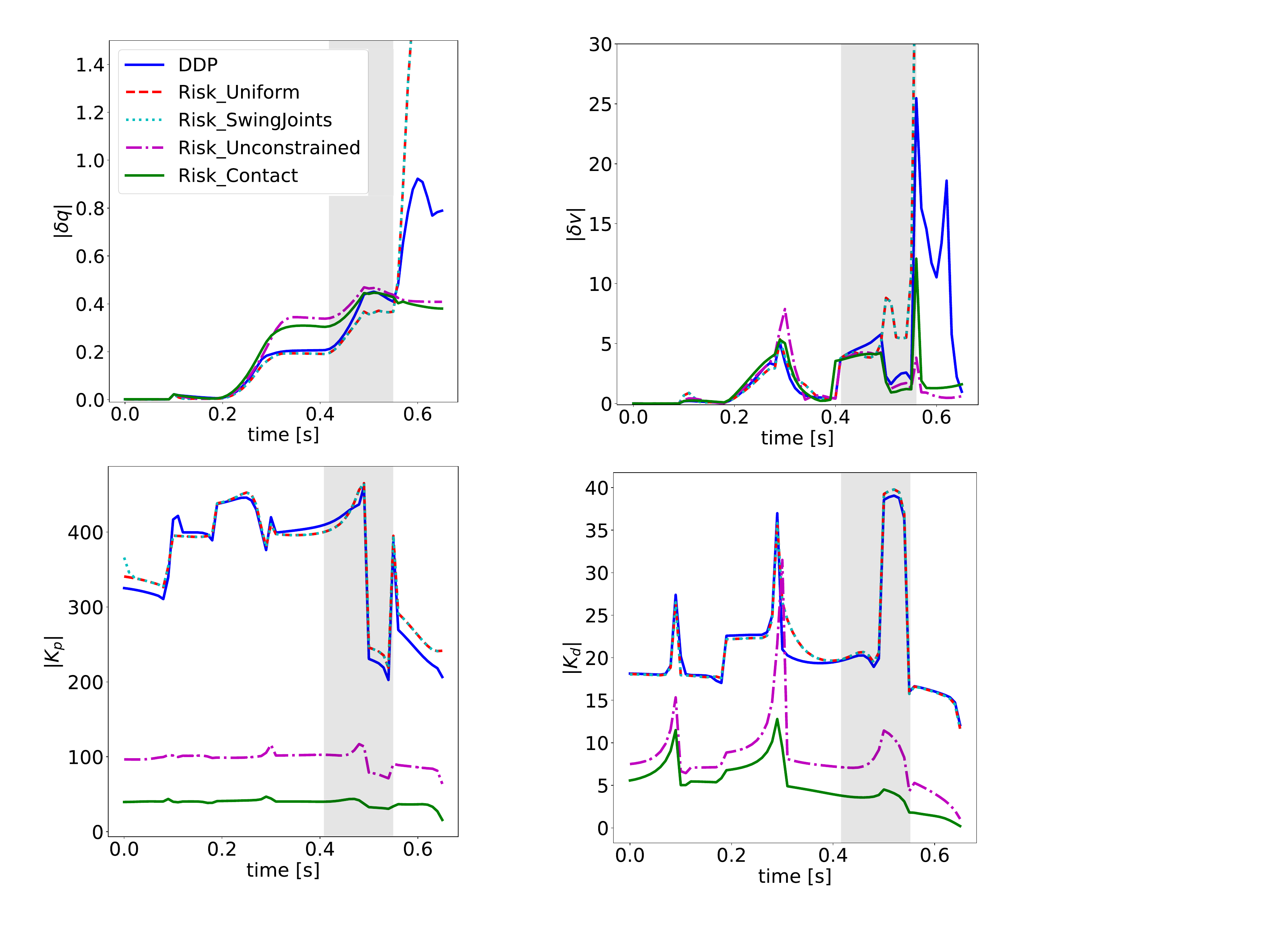}
\label{fig:exp01_kd}}
\caption{Tracking error and feedback norms for uncertainty models and DDP. The grey zone corresponds to the time between the unexpected contact and the planned one.}
\label{fig:exp01_feedback_tracking}
\vspace{-0.5cm}
\end{figure}

The first uncertainty model, \DIFdelbegin \textbf{\DIFdel{Risk-Uniform}}
\DIFdelend \DIFaddbegin {\em \DIFadd{Risk-Uniform}}\DIFaddend , is simply a diagonal matrix with equal variance on all of its entries. In the second uncertainty model, \DIFdelbegin \textbf{\DIFdel{Risk-SwingJoints}}
\DIFdelend \DIFaddbegin {\em \DIFadd{Risk-SwingJoints}}\DIFaddend , the variance terms on the joints of the swinging foot are increased. In the third model, \DIFdelbegin \textbf{\DIFdel{Risk-Unconstrained}}
\DIFdelend \DIFaddbegin {\em \DIFadd{Risk-Unconstrained}}\DIFaddend , we add a contact noise term similar to~\eqref{Eq:contact_uncertainty_model} without using the nullspace projection due to the active contacts. The last model, \DIFdelbegin \textbf{\DIFdel{Risk-Contact}}
\DIFdelend \DIFaddbegin {\em \DIFadd{Risk-Contact}}\DIFaddend , includes the projection of the swing contact uncertainty into the null space of the active contacts~\eqref{Eq:contact_uncertainty_model_constrained}.

\begin{figure}[h]
\DIFdelbeginFL 
\DIFdelendFL 
    \centering
    \captionsetup{justification=centering}
  \includegraphics[width=.7\linewidth]{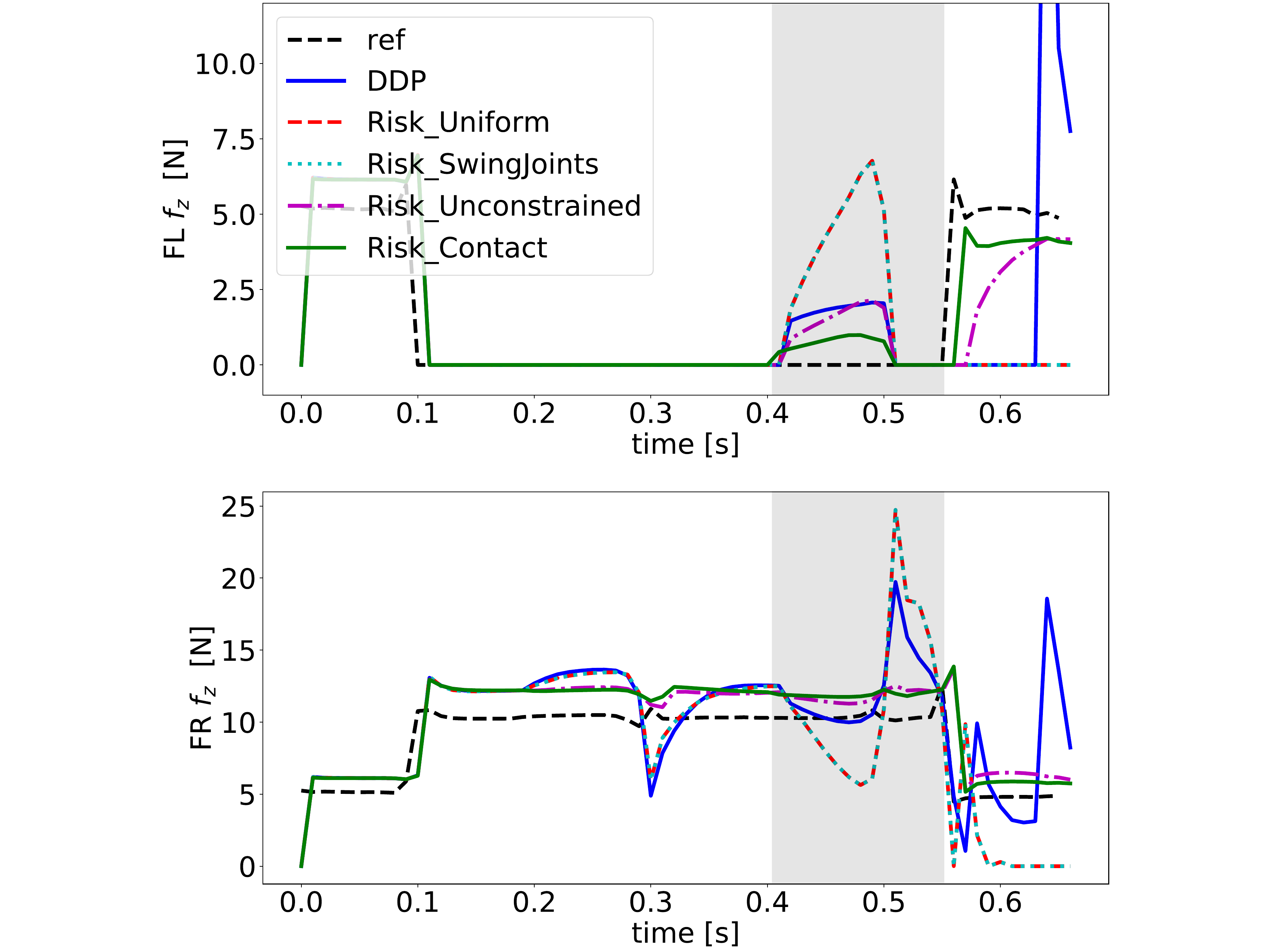}
  \caption{Front feet normal forces for different uncertainty models and DDP. 
  }
  \label{fig:exp01_contact_force}
  \DIFdelbeginFL 

{
\DIFdelFL{Part of the feedback gains mapping base errors to joint torques at the time of unexpected impact t = }
}
\DIFdelendFL \vspace{-0.3cm}
\end{figure}

Figure~\ref{fig:exp01_feedback_tracking} compares the tracking behavior of each control scheme. The state error is divided into the error in the configuration $\delta q$ ( Fig.~\ref{fig:exp01_q_error}) and the error in the velocity tracking $\delta v$ (Fig.~\ref{fig:exp01_v_error}). Similarly the optimized feedback gains are divided into the feedback gains associated to the configuration $K_p$ and the feedback gains from the velocity $K_d$. Their norms are shown in Fig.~\ref{fig:exp01_kp} and Fig.~\ref{fig:exp01_kd} respectively. The vector norm $\Vert v \Vert =  \sqrt{v^T.v}$ is used to compute the state error norms whereas the Frobenius Norm $\Vert M \Vert_F = \sqrt{Trace\left(M.M^T\right)}$ is used to compute the feedback norms. 

We notice in Fig.~\ref{fig:exp01_feedback_tracking} that the risk sensitive control with Risk-Uniform and Risk-SwingJoint noise models diverge after the unexpected impact. DDP also diverges but accumulates less error than these two risk sensitive schemes. All three controllers have high gain norms at the time of impact $\vert Kp\vert \approx 400$ and a stiffness to damping ratio at $ \sfrac{\vert Kp\vert}{\vert Kd\vert} \approx 20$. 
It is important to note that such gains are too high and would not be usable on the real robot\DIFaddbegin \DIFadd{~}\DIFaddend \cite{grimminger2020open}.
The maximum reduction in the overall stiffness and damping is obtained by using the Contact uncertainty model where $\sfrac{\vert Kp_{DDP}\vert}{\vert Kp_{Contact}\vert} \approx 10$ and  $\sfrac{\vert Kd_{DDP}\vert}{\vert Kd_{Contact}\vert} \approx 4$. With this reduction in the impedance magnitudes, a less accurate tracking is observed for both Unconstrained and Contact earlier in time $t=\SI{0.3}{s}$. However, thanks to the less aggressive feedback gains, the robot can handle the unpredicted impact. Importantly, these reduced gains fit well within ranges acceptable for execution on the real robot.

\begin{figure}[h]
\centering
\captionsetup{justification=centering}
\subfloat[][Base Stiffness]{
\includegraphics[width=0.75\linewidth]{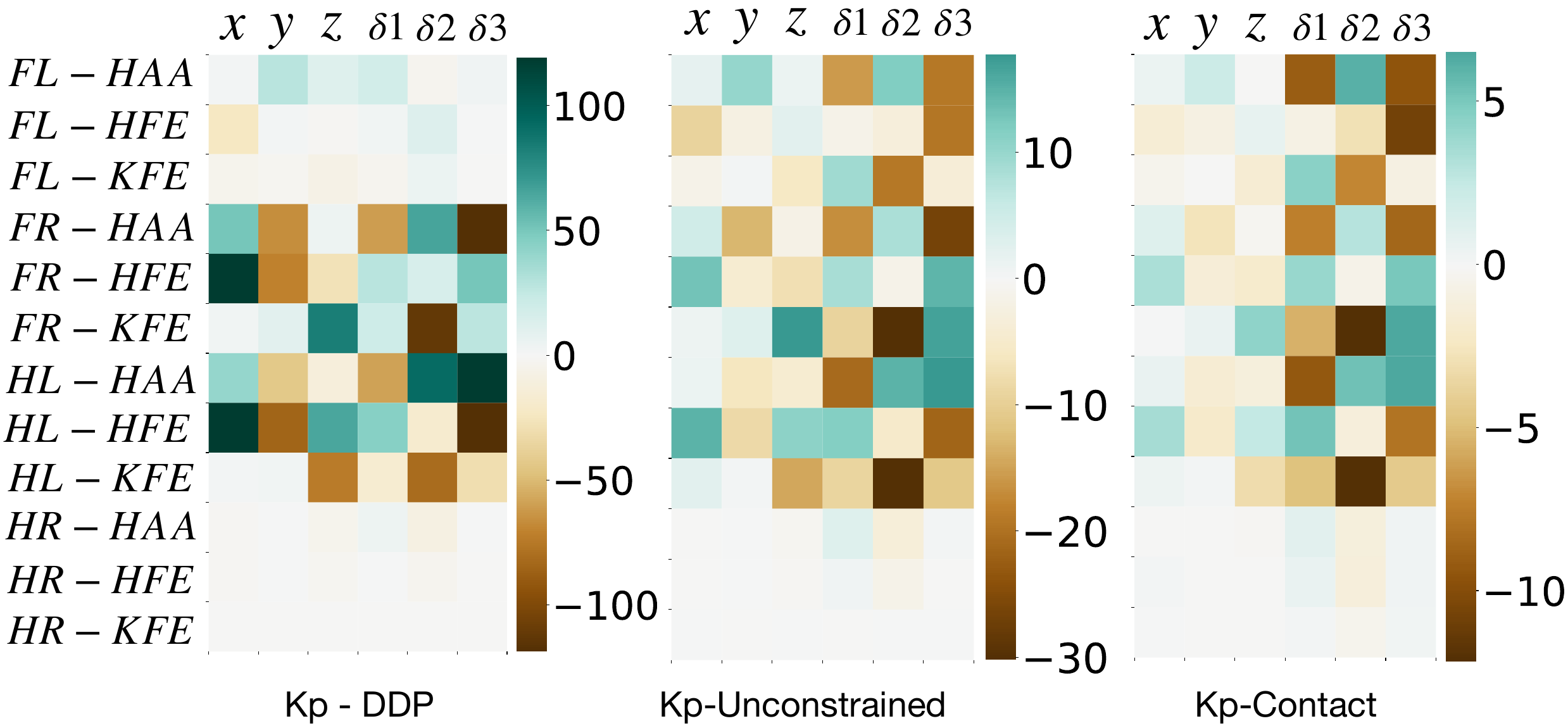}
\label{fig:exp01_base_kp}} \\
\subfloat[][Base Damping]{
\includegraphics[width=0.75\linewidth]{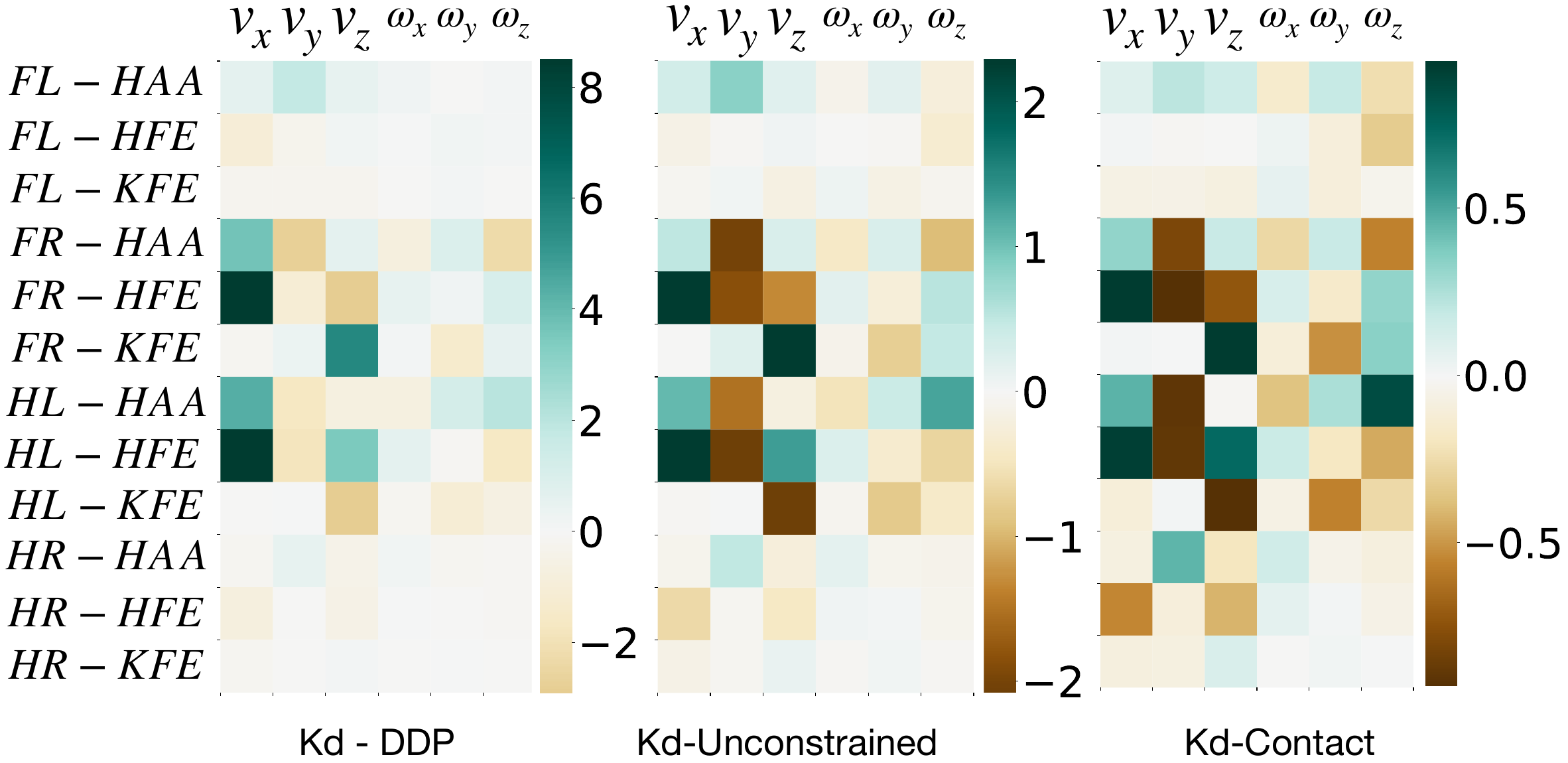}
\label{fig:exp01_base_kd}}
\caption{Part of the feedback gains mapping base errors to joint torques at the time of unexpected impact t = \SI{0.43}{s}}
\label{fig:exp01_feedback_base}
\vspace{-0.2cm}
\end{figure}

The effect of the controllers is clearer when looking at the normal contact forces (Fig.~\ref{fig:exp01_contact_force}) where the impact force on the swinging leg $FL$ after $t=\SI{0.43}{s}$ is lowest for Risk-Contact. 
Additionally, the impact force propagates its effect to the front right leg for DDP, Risk-Uniform and Risk-SwingJoints. Whereas for the case where the uncertainty about the contact is included (Risk-Unconstrained and Risk-Contact), we see that the robot absorbs the impact force and we do not notice any propagation of the disturbance to the other feet. 

\begin{figure}[h]
\vspace{0.2cm}
\centering
\captionsetup{justification=centering}
\subfloat[][Base Stiffness]{
\includegraphics[width=0.75\linewidth]{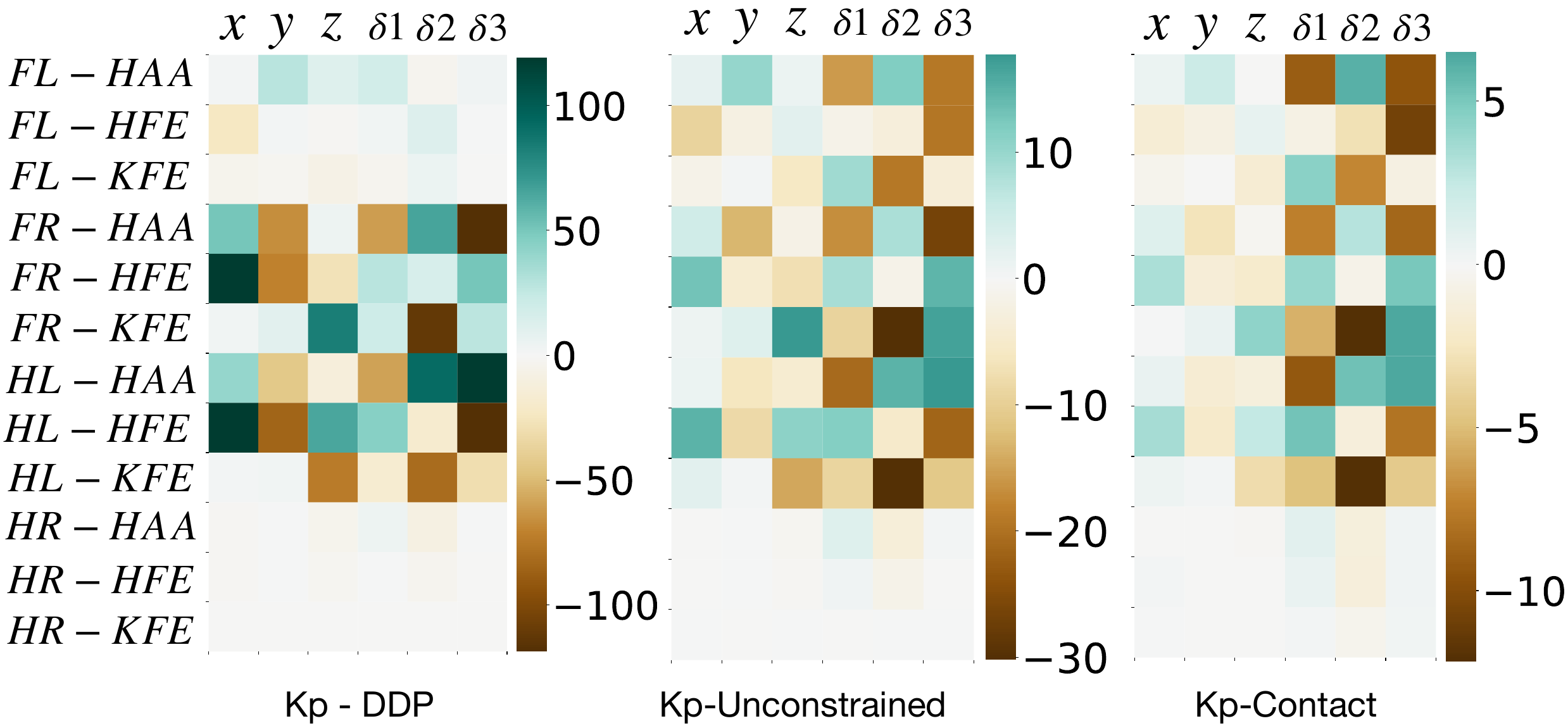}
\label{fig:exp01_base_kp}} \\
\subfloat[][Base Damping]{
\includegraphics[width=0.75\linewidth]{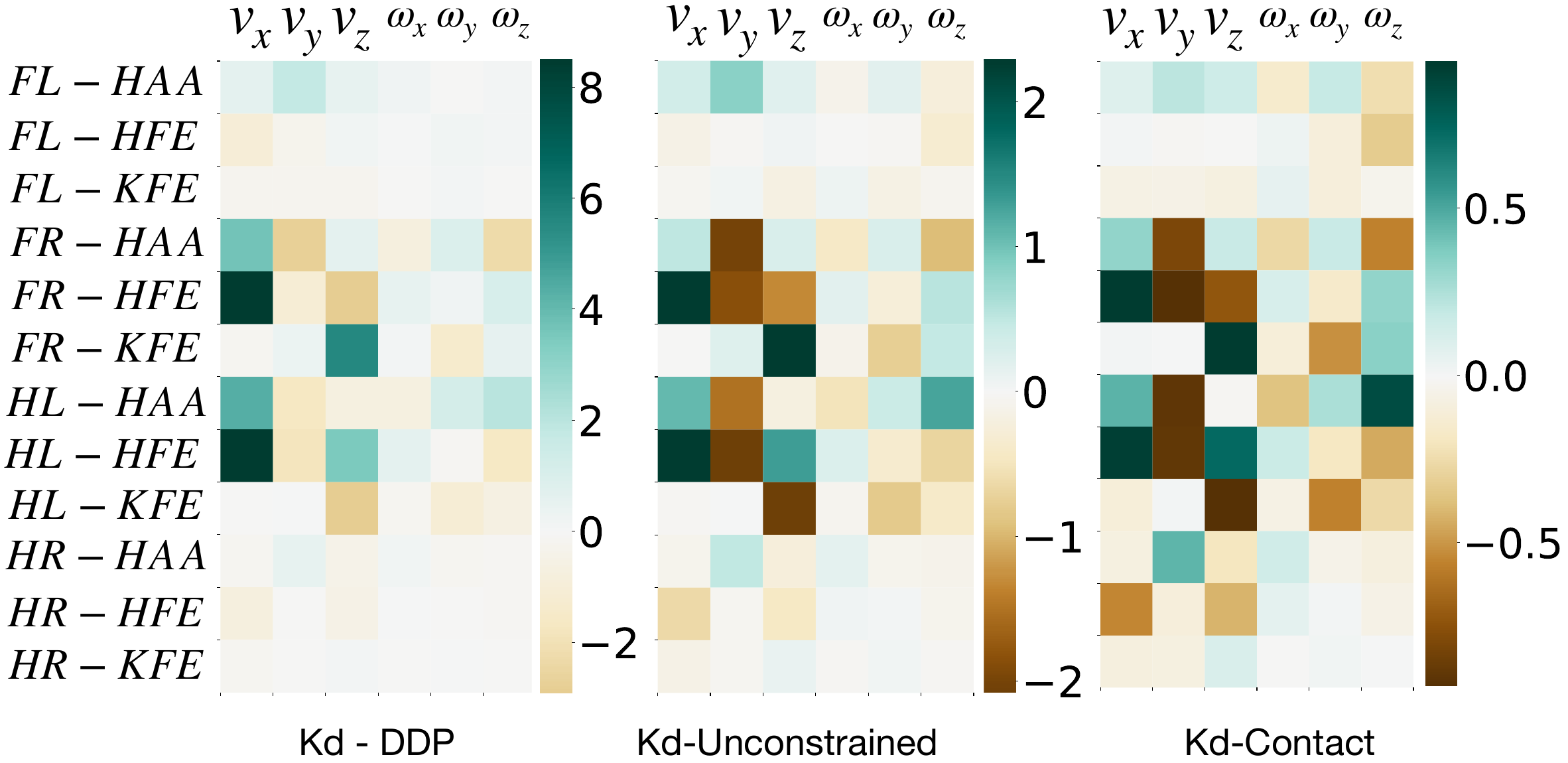}
\label{fig:exp01_base_kd}}
\caption{Part of the feedback gains mapping base errors to joint torques at the time of unexpected impact t = \SI{0.43}{s}}
\label{fig:exp01_feedback_base}
\vspace{-0.3cm}
\end{figure}

Inspecting the structure of the feedback matrices computed at the time of impact can shed light on these differences in behavior.
The blocks of the feedback matrix that map the base states to the control are depicted in Fig.~\ref{fig:exp01_feedback_base}. Risk Sensitive control changes the structure of the gain matrices, where the base error is modulated mainly through the feet on the ground namely FR, HL and HR, which is expected.  The advantage of Risk-Contact is observed in the portion of the feedback matrix that maps the joint errors to the control commands (Fig.~\ref{fig:exp01_feedback_joint}). DDP has aggressive Kp gain to track the motion of the swinging foot FL, that is the first three diagonal elements of Kp-DDP relative to the remaining diagonal elements corresponding to the remaining joints. Unlike Kp-DDP and Kp-Unconstrained, Kp-Contact has lower gains on the joints of FL relative to the joints of the support feet FR and HL. This allows the swing foot to behave as a soft spring relative to the support feet which explains the significantly lower impact force observed and the good tracking of the contact forces on the other feet. 
These results underline the importance of choosing appropriate noise models and supports the choice of the Risk-Contact noise model.

\subsection{Stiffness vs. Damping \& Impact Forces}
We now only consider the Risk-Contact noise model. This experiment discusses how the introduced model results in an improvement of the performance of an extremely dynamic motion, a jump of a total height of $\SI{0.5}{m}$. Halfway through the flight phase, a block of $\SI{3}{cm}$ height is placed on the floor. \DIFdelbegin \DIFdel{We model this uncertainty by increasing the contact-specific measurement noise.
}\DIFdelend 

\begin{figure}[h]
\centering
\captionsetup{justification=centering}
\subfloat[][Base $\vert \delta q \vert$]{
\includegraphics[width=0.45\linewidth]{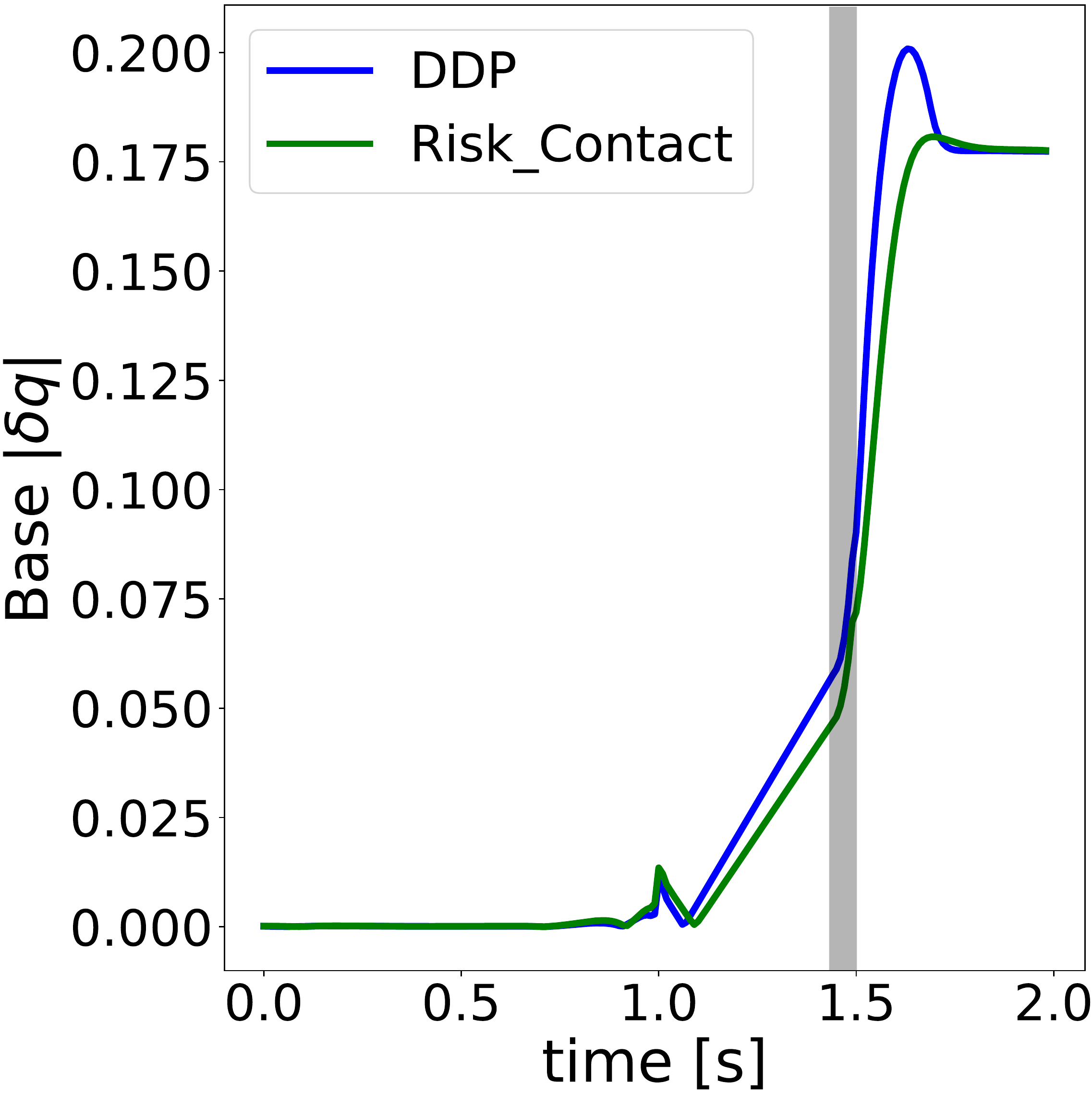}
\label{fig:exp02_base_qerror}}
\subfloat[][Joint $\vert \delta q \vert$]{
\includegraphics[width=0.45\linewidth]{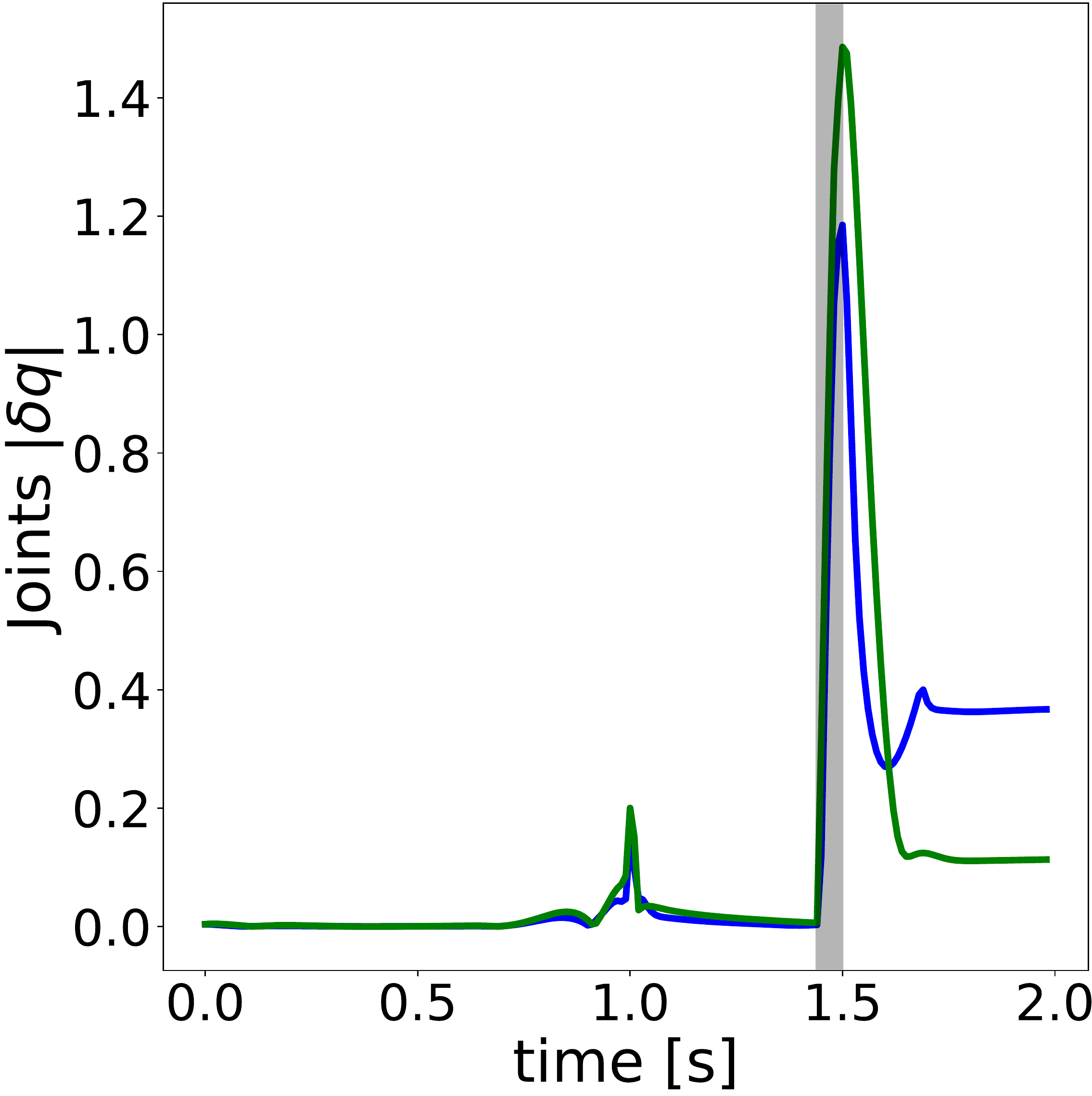}
\label{fig:exp02_joint_qerror}} \\
\subfloat[][Base $\vert K_p \vert$]{
\includegraphics[width=0.45\linewidth]{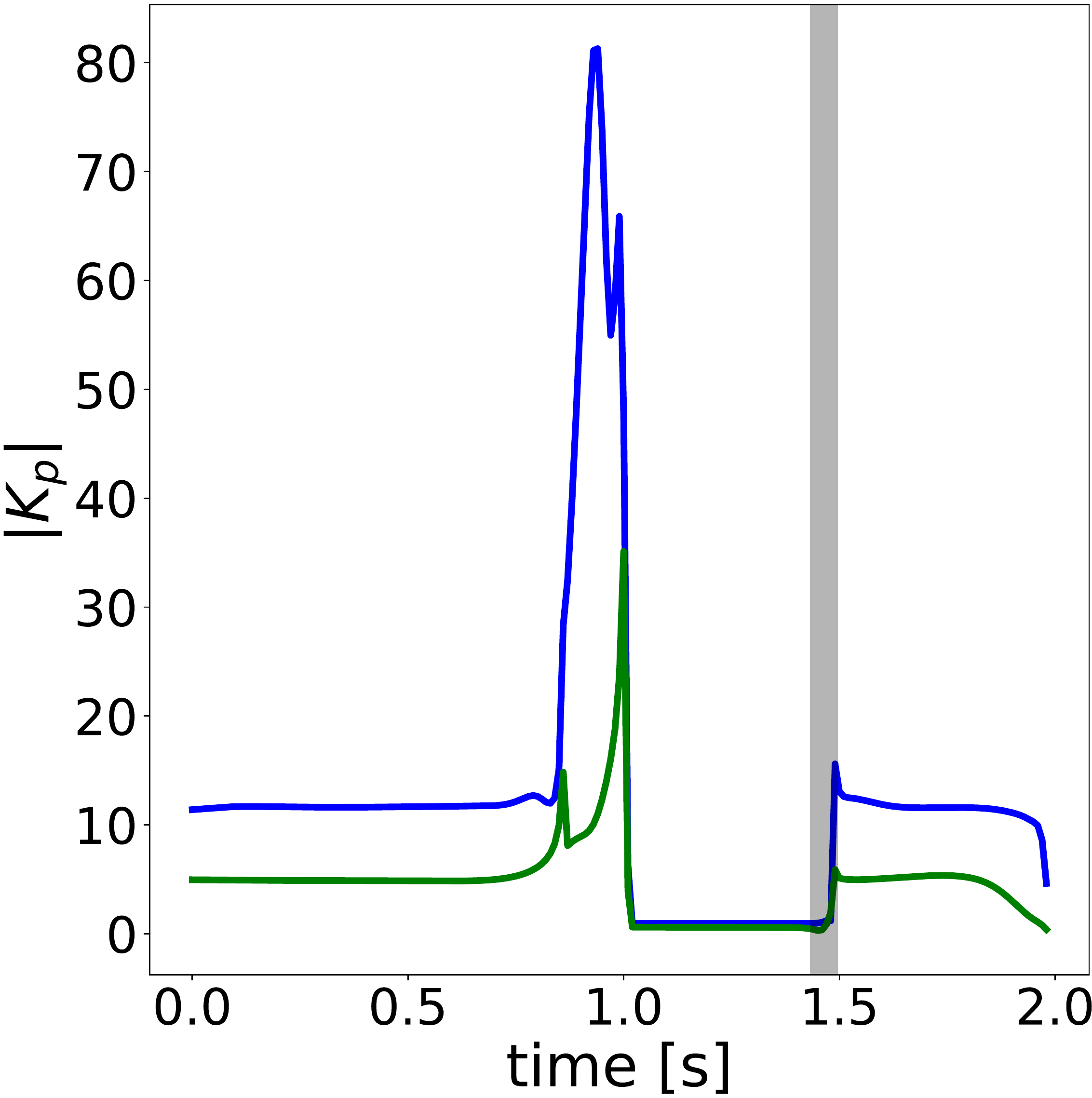}
\label{fig:exp02_base_kp}}
\subfloat[][Joint $\vert K_p \vert$]{
\includegraphics[width=0.45\linewidth]{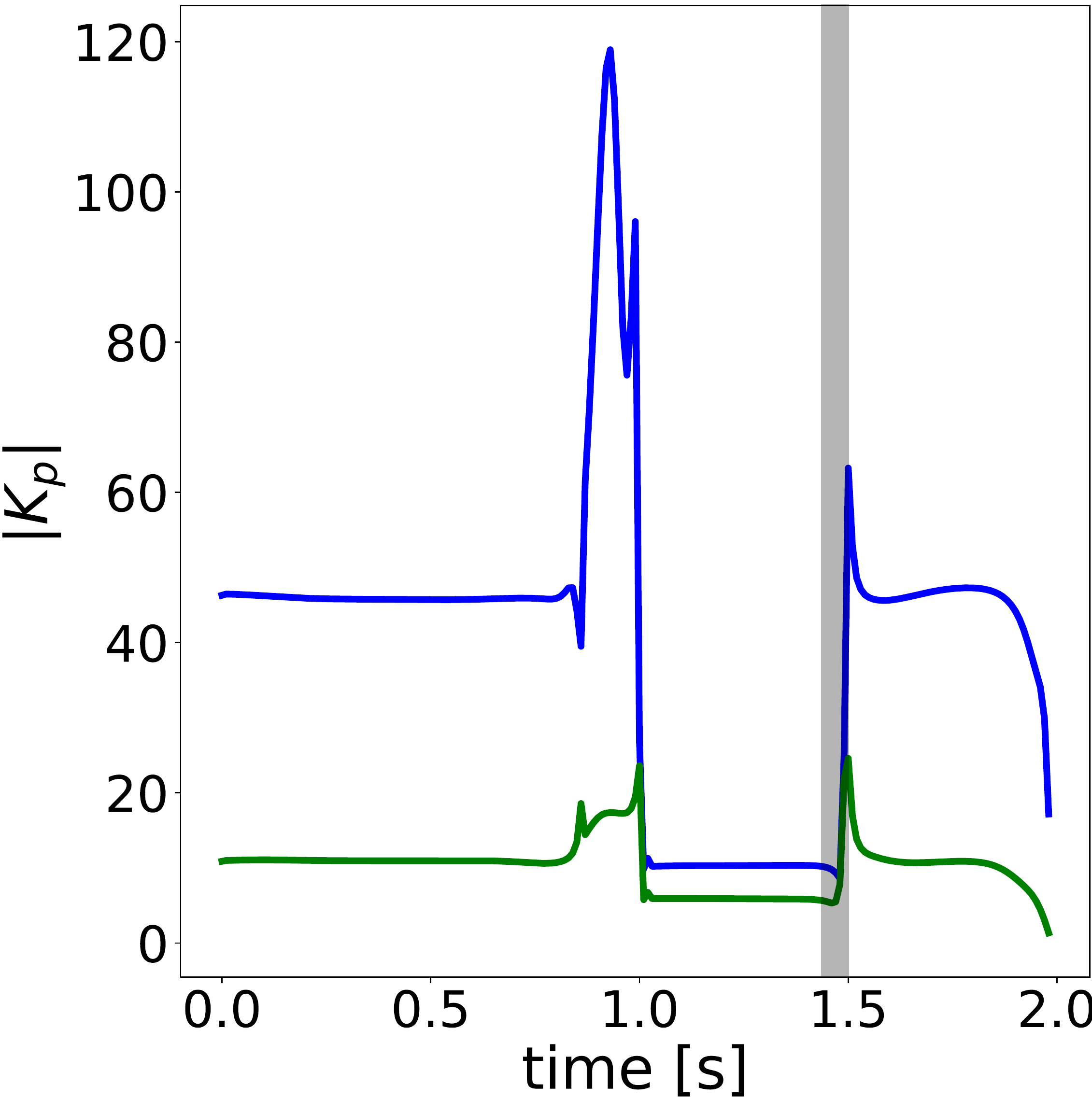}
\label{fig:exp02_joint_kp}}
\caption{Stiffness norm and position errors.}
\label{fig:exp02_position_stiffness}
\vspace{-0.4cm}
\end{figure}

\begin{figure}[h]
\vspace{0.2cm}
\centering
\captionsetup{justification=centering}
\subfloat[][Base $\vert \delta v \vert$]{
\includegraphics[width=0.45\linewidth]{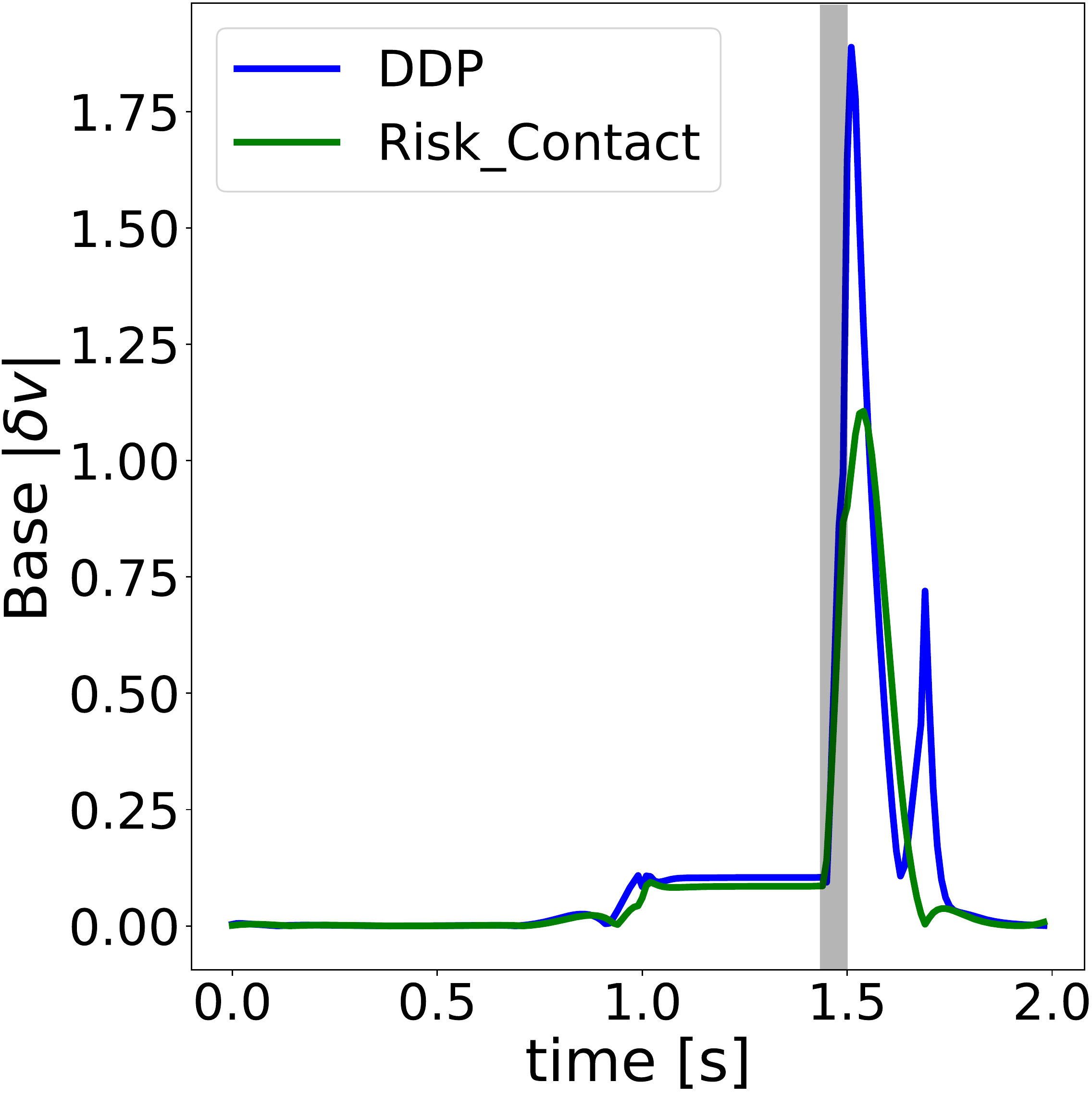}
\label{fig:exp02_base_verror}}
\subfloat[][Joint $\vert \delta v \vert$]{
\includegraphics[width=0.45\linewidth]{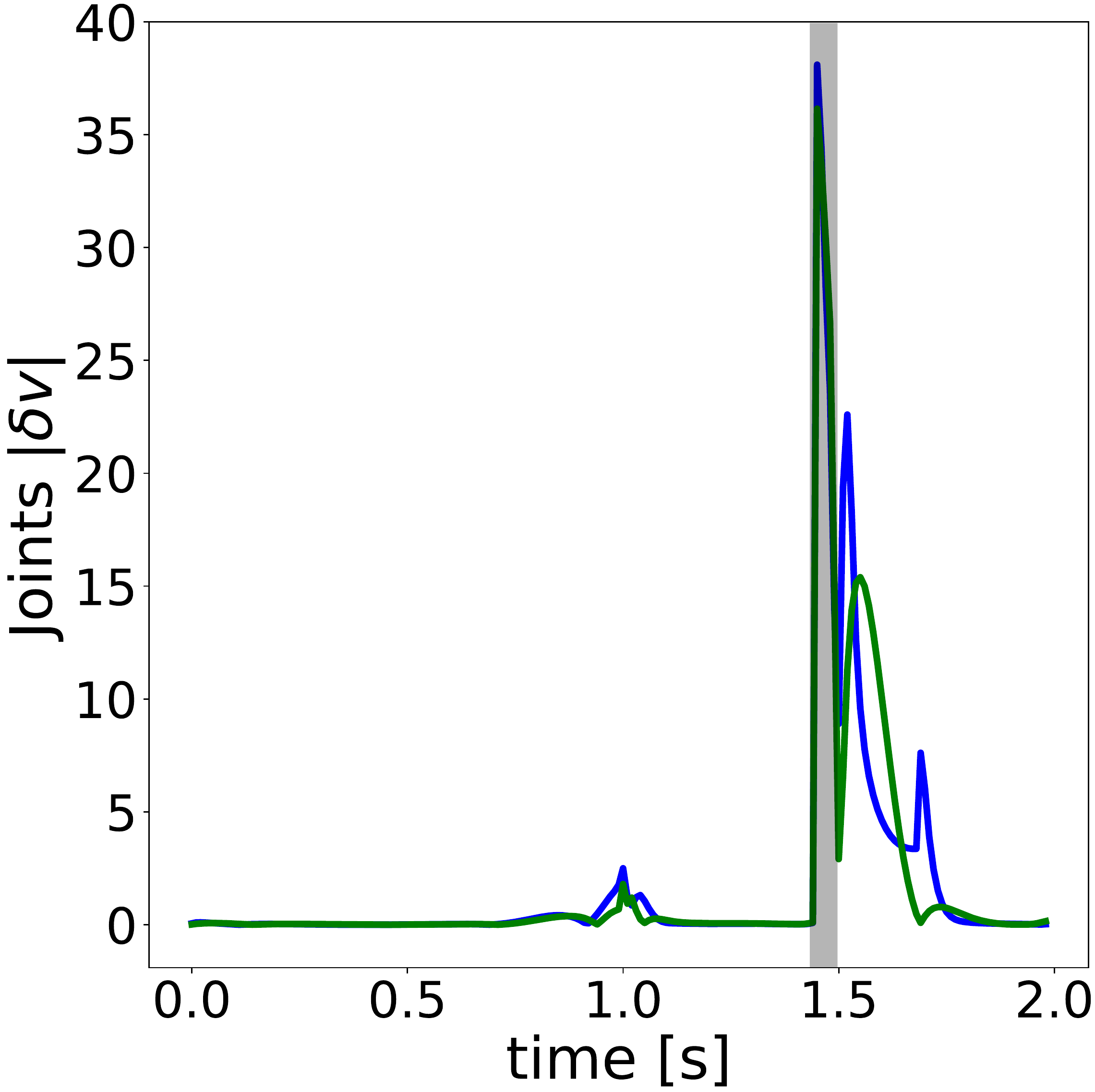}
\label{fig:exp02_joint_verror}} \\
\subfloat[][Base $\vert K_d \vert$]{
\includegraphics[width=0.45\linewidth]{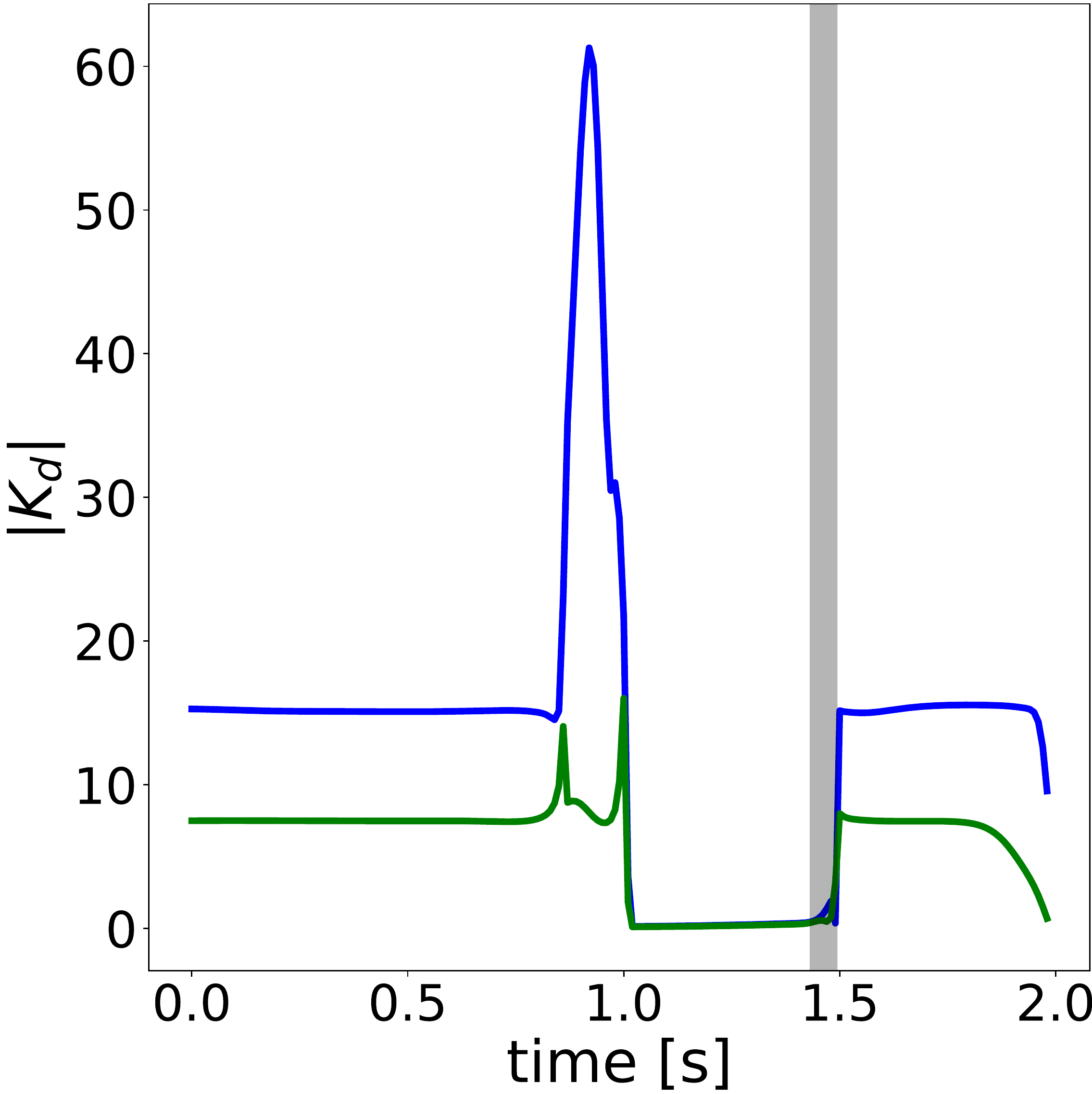}
\label{fig:exp02_base_kd}}
\subfloat[][Joint $\vert K_d \vert$]{
\includegraphics[width=0.45\linewidth]{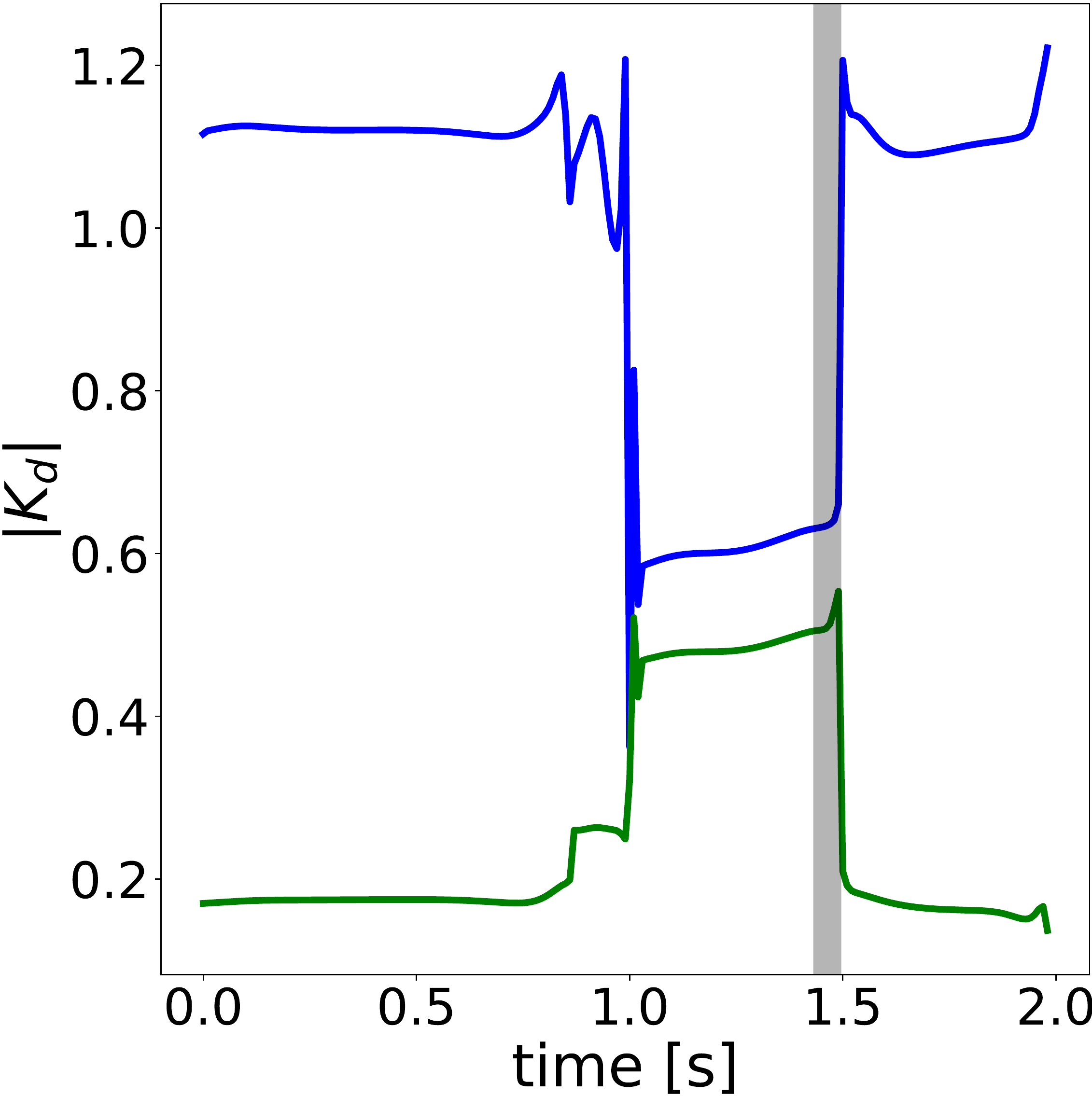}
\label{fig:exp02_joint_kd}} 
\caption{Damping and Velocity Errors}
\label{fig:exp02_velocity_damping}
\end{figure}

\begin{figure}[h]
\centering
\captionsetup{justification=centering}
\includegraphics[width=0.75\linewidth]{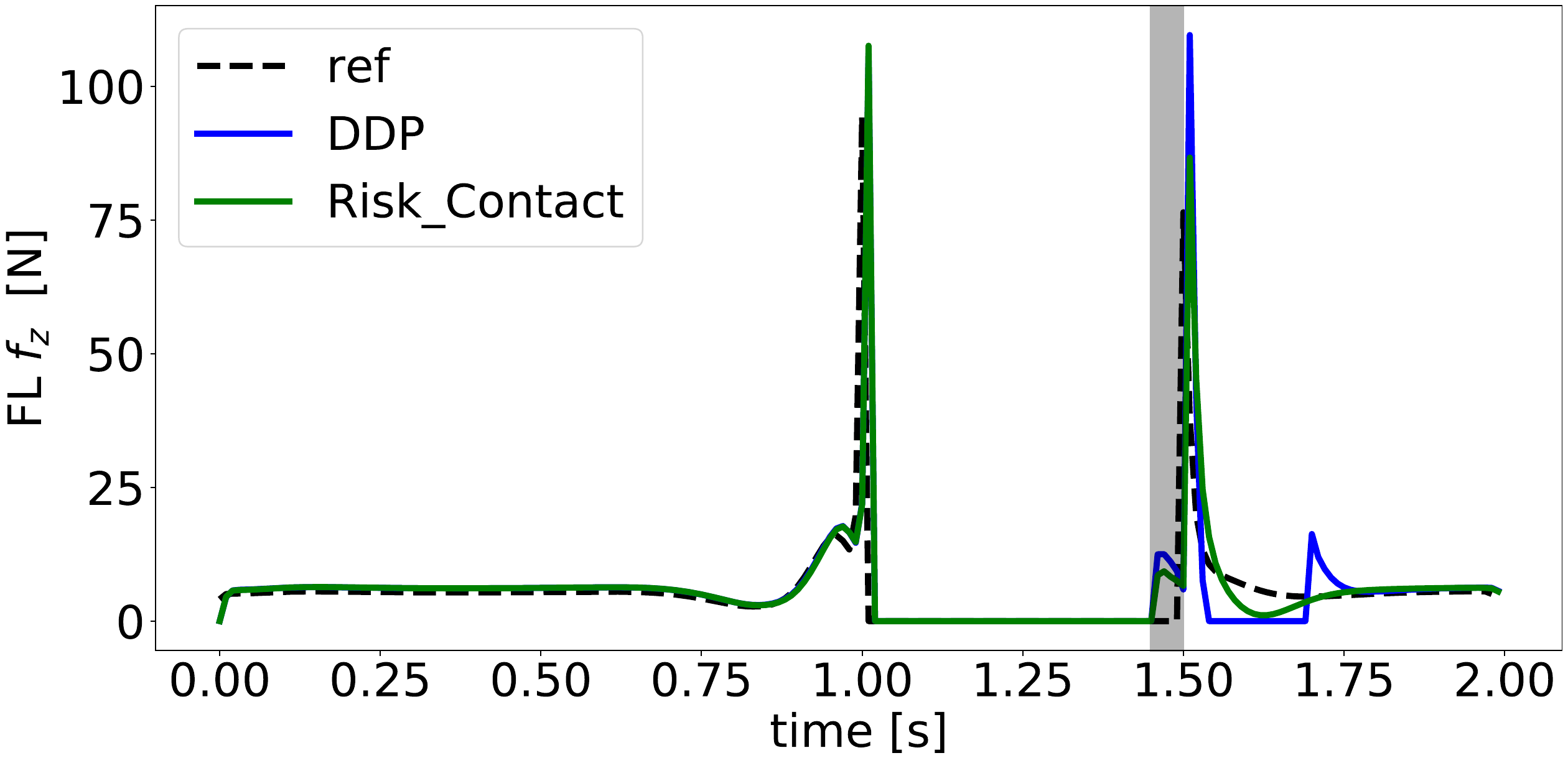}
\caption{FL Normal Contact Forces}
\label{fig:exp02_normal_forces} 
\vspace{-0.5cm}
\end{figure}

\begin{figure}[h]
\vspace{0.2cm}
\centering
\captionsetup{justification=centering}
\subfloat[][Base Height]{
\includegraphics[width=0.7\linewidth]{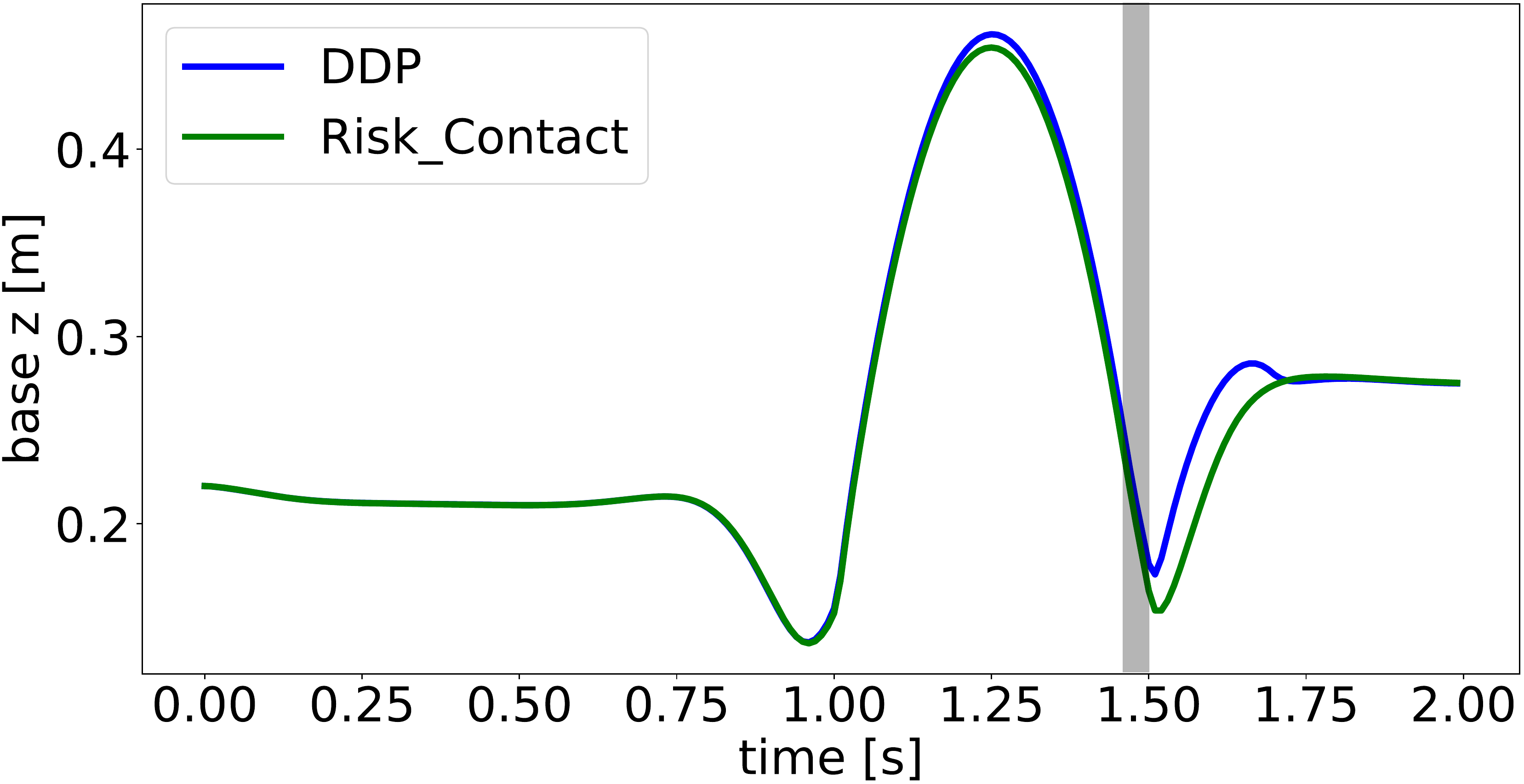}
\label{fig:exp02_base_height}} \\
\subfloat[][Front Left Foot Height]{
\includegraphics[width=0.73\linewidth]{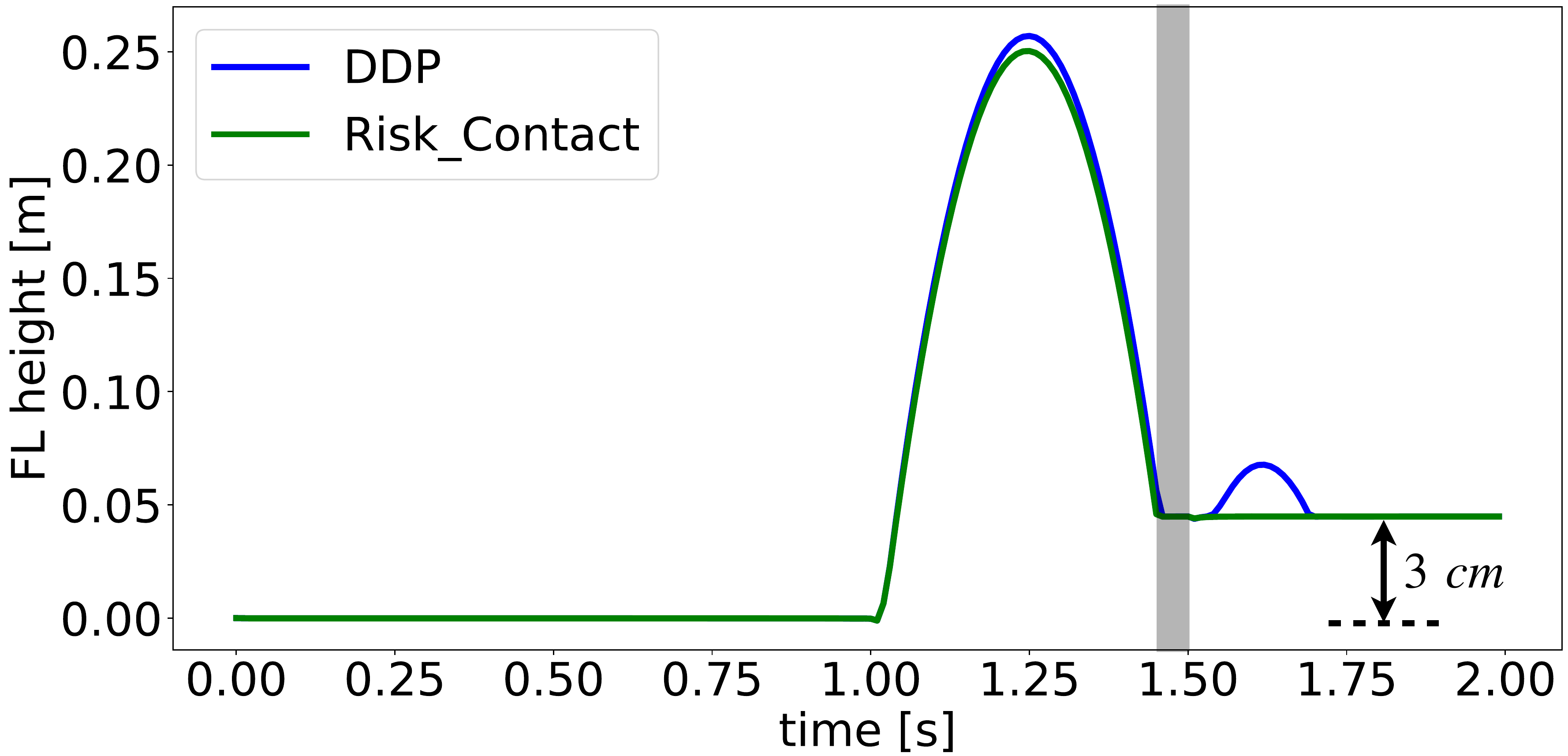}
\label{fig:exp02_contact_height}} 
\caption{Base and Foot Height Comparison}
\label{fig:exp02_jump_contact}
\vspace{-0.5cm}
\end{figure}

Both $DDP$ and Risk-Contact feedback policies achieve the desired jump height with a slightly better performance for the DDP controller. However, inspecting the tracking and impedance profiles reveals a more natural behavior emerging in the Risk Sensitive case. During the flight phase between $t=\SI{1}{s}$ and  \DIFdelbegin \DIFdel{$t={1.5}{s}$ }\DIFdelend \DIFaddbegin \DIFadd{$t=\SI{1.5}{s}$ }\DIFaddend both controllers exhibit a significant decrease in both the base and joint stiffness as shown in Fig.~\ref{fig:exp02_position_stiffness}. For the damping portion of the feedback matrix, the base damping also shows a decrease during the flight phase. Remarkably, we notice a stark difference in the damping modulation of the joints. DDP significantly decreases damping on the joints while Risk Sensitive significantly increases it. This substantial increase in the joint damping allows to more quickly
absorb the unexpected impact, i.e. an abrupt change in the velocity of the feet.
As a result, we notice $20\%$ lower impact forces (Fig.~\ref{fig:exp02_normal_forces}), which in turn avoids the bouncing behavior observed with the DDP controller (Fig.~\ref{fig:exp02_jump_contact}). This comes however at the cost of larger deviations for the joint positions.
When measuring the stiffness to damping ratio of both the base and the joints, we notice that during active contact phases the base stiffness to damping ratio is relatively the same for DDP and risk sensitive. However, for the joints, during the support phase, the stiffness to damping ratio of risk sensitive is 1.5 times higher than that of DDP, explaining the good tracking with lower overall gain profiles. 

\begin{figure*}[h]
\vspace{0.2cm}
\centering
\captionsetup{justification=centering}
\includegraphics[width=.7\linewidth]{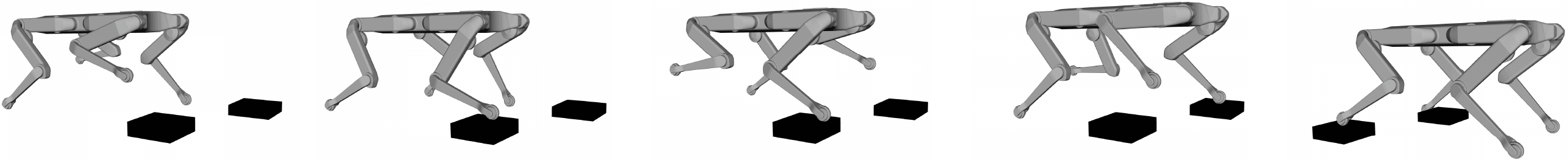}
\caption{Snapshots of the trotting gait and the contact disturbances.}
\label{fig:exp03_banner}
\vspace{-0.3cm}
\end{figure*}

\subsection{Trade-off between Stability \& Accuracy}
In this experiment, we study the capabilities of our controller when trotting on unknown terrain. Both DDP and risk sensitive policies are optimized to track a trotting gait including a total of 14 contact switches. During simulations, we introduce unexpected blocks at contact locations (Fig.~\ref{fig:exp03_banner}) in order to study the policies robustness. An execution is successful if the robot manages to reach the desired terminal base configuration and velocity without falling.

\begin{table}[h]
\centering
\captionsetup{justification=centering}
\begin{tabular}{c|c|c|c|c|}
\cline{2-5}
                                            & \multicolumn{2}{c|}{Experiment 1} & \multicolumn{2}{c|}{Experiment 2} \\ \hline
\multicolumn{1}{|c|}{\# of Simulations}     & \multicolumn{2}{c|}{500}           & \multicolumn{2}{c|}{500}           \\ \hline
\multicolumn{1}{|c|}{Maximum \# of blocks}  & \multicolumn{2}{c|}{14}            & \multicolumn{2}{c|}{14}            \\ \hline
\multicolumn{1}{|c|}{Maximum block height} & \multicolumn{2}{c|}{45 mm} & \multicolumn{2}{c|}{34 mm} \\ \hline
\multicolumn{1}{|c|}{\% of leg length}     & \multicolumn{2}{c|}{14 \%} & \multicolumn{2}{c|}{11 \%} \\ \hline
\multicolumn{1}{|c|}{Method}                & DDP     & \textbf{Risk}  & DDP     & \textbf{Risk}  \\ \hline
\multicolumn{1}{|c|}{\# of Successful Sim.} & 237     & \textbf{325}             & 358     & \textbf{458}             \\ \hline
\multicolumn{1}{|c|}{\% of Successful Sim.} & 47.4\%  & \textbf{65\%}            & 72.6\%  & \textbf{91.6\%}          \\ \hline
\end{tabular}
\caption{Parameters and results of the trotting experiments}
\label{tab:my-table}
\vspace{-0.4cm}
\end{table}
\begin{figure}[h]
\centering
\captionsetup{justification=centering}
\subfloat[][Position error norm]{
\includegraphics[width=0.45\linewidth]{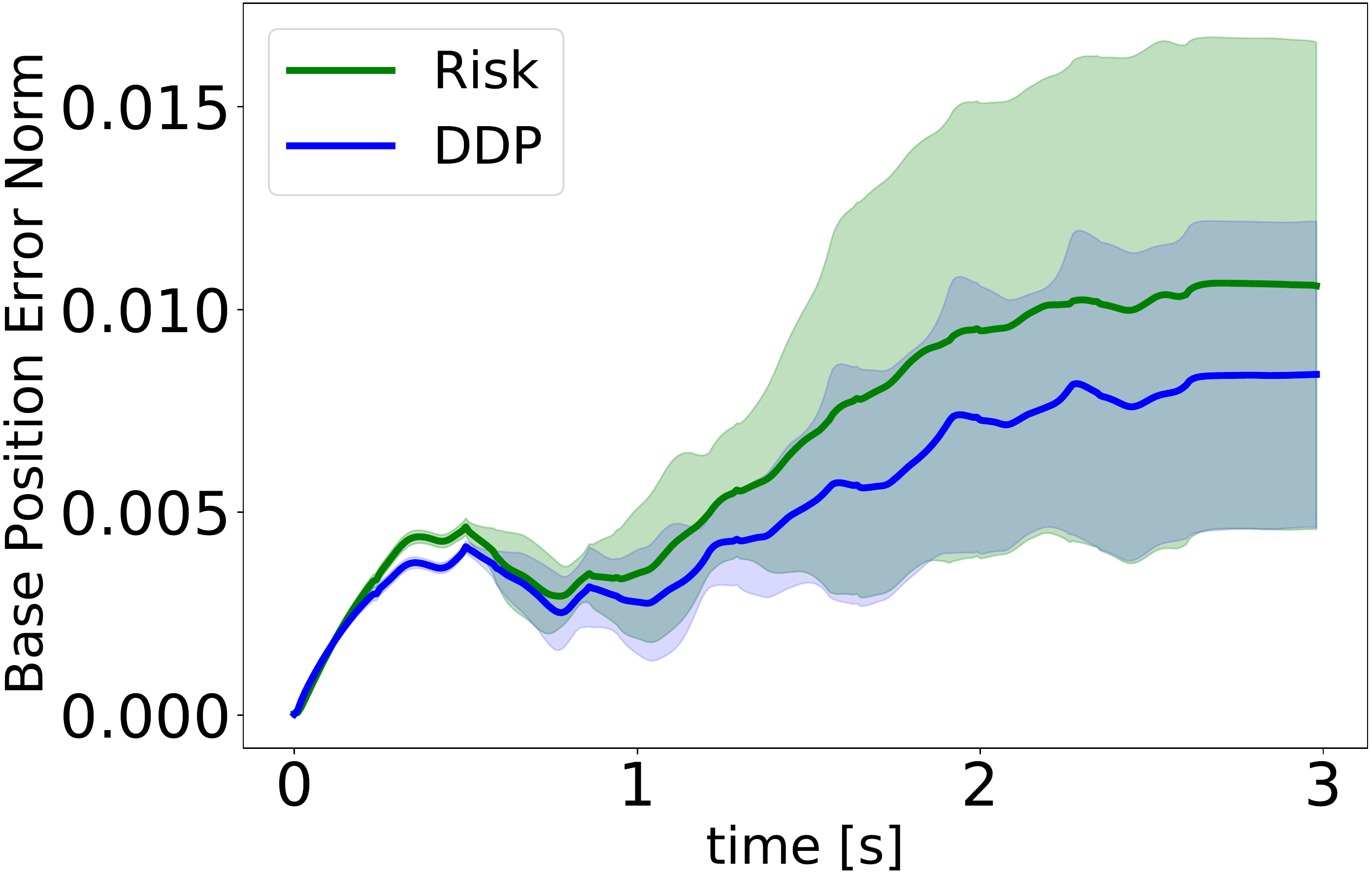}
\label{fig:exp03_position_error}} 
\subfloat[][Orientation error norm]{
\includegraphics[width=0.451\linewidth]{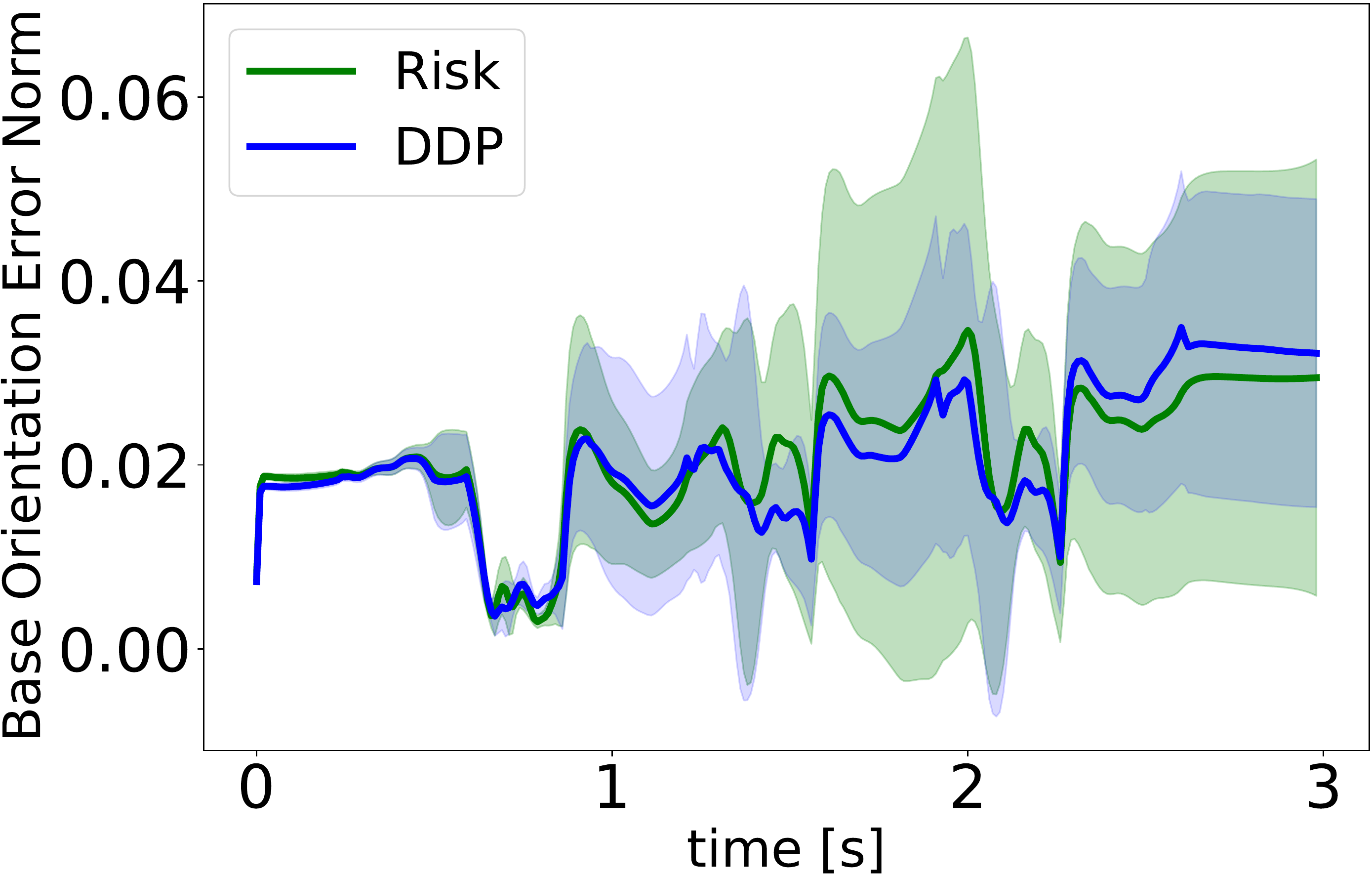}
\label{fig:exp03_orientation_error}} 
\caption{Trotting Base Tracking Error Norms Statistics}
\label{fig:exp03_error_norms}
\vspace{-0.6cm}
\end{figure}
We conduct 1000 simulations, divided into two batches. In the first batch, the contact disturbances
are sampled to represent up to $14 \%$ of the total leg length, which is quite aggressive. In the second batch, the maximum contact variation is $11 \%$ of the leg length. The results are summarized in Table \ref{tab:my-table}.
In both cases, we observe a significant increase in trotting performance when using the risk sensitive controller. Moreover, the magnitude of the gains of the DDP controller are outside of the range of gains admissible on the real robot while the risk sensitive controller gains are not \DIFaddbegin \DIFadd{(based on our preliminary investigations on the real robot)}\DIFaddend .
This result demonstrates \DIFdelbegin \DIFdel{the improved }\DIFdelend \DIFaddbegin \DIFadd{that the risk-sensitive controller effectively improves the }\DIFaddend robustness to contact uncertainties\DIFdelbegin \DIFdel{provided by the risk-sensitive controller}\DIFdelend .

For successful executions, we show the distribution of base position \DIFdelbegin \DIFdel{and velocity }\DIFdelend tracking errors in Fig.~\ref{fig:exp03_error_norms}. When successful, the risk sensitive control policy finishes the trotting gait with larger deviations in the base configuration \DIFdelbegin \DIFdel{while damping large velocity disturbances to stabilize the robot}\DIFdelend \DIFaddbegin \DIFadd{to generate higher number of stable execution}\DIFaddend , underlying the trade-off between accurate tracking and robustness to contact uncertainty.

\DIFaddend \section{Conclusion}\label{sec:conclusion}
In this paper, we extended the idea of risk sensitive optimal control
with measurement uncertainty to legged locomotion problems.
We showed the importance of the choice of the noise models to generate
meaningful impedance modulation patterns.
Through extensive simulations, we demonstrated that our approach
can generate stiffness and damping profiles that lead to better responses
in face of hard impacts and contact uncertainty when compared to 
typical DDP algorithms.
\DIFdelbegin \DIFdel{Moreover, the }\DIFdelend \DIFaddbegin \DIFadd{The }\DIFaddend computed gains are significantly smaller\DIFdelbegin \DIFdel{than the ones computed through DDP}\DIFdelend , and within ranges that are realistically executable
on the real robot.
Our approach provides a systematic approach to automatically compute
optimal impedance modulations, at the same computational cost as modern
DDP algorithms. Future work will include \DIFdelbegin \DIFdel{experiments on real robots and
efforts to extend the approach }\DIFdelend \DIFaddbegin \DIFadd{real robot experiments}.
\vspace{-0.1cm}
\bibliographystyle{IEEEtran}
\bibliography{references}

\end{document}